\documentclass[nohyperref]{article}

\usepackage{microtype}      
\usepackage{graphicx}
\usepackage{subfigure}
\usepackage{booktabs}       

\usepackage{hyperref}       

\usepackage[accepted]{icml2022}

\usepackage{amsmath}
\usepackage{amssymb}
\usepackage{mathtools}
\usepackage{amsthm}
\usepackage{amsfonts}       
\usepackage{nicefrac}       
\usepackage{enumitem}
\usepackage[para]{footmisc}

\usepackage{url}            

\usepackage{natbib} 
\icmltitlerunning{Data augmentation for efficient learning from parametric experts}

\begin{document}

\twocolumn[
\icmltitle{Data augmentation for efficient learning from parametric experts}

\icmlsetsymbol{equal}{*}

\begin{icmlauthorlist}
\icmlauthor{Alexandre Galashov}{DM}
\icmlauthor{Josh Merel}{DM}
\icmlauthor{Nicolas Heess}{DM}
\end{icmlauthorlist}

\icmlaffiliation{DM}{DeepMind}
\icmlcorrespondingauthor{Alexandre Galashov}{galashov.alexandr@gmail.com}

\icmlkeywords{Machine Learning, ICML}

\vskip 0.3in
]



\printAffiliationsAndNotice{\icmlEqualContribution} 

\begin{abstract}
We present a simple, yet effective data-augmentation technique to enable data-efficient learning from parametric experts for reinforcement and imitation learning. We focus on what we call the \textit{policy cloning} setting, in which we use online or offline queries of an expert or expert policy to inform the behavior of a student policy. This setting arises naturally in a number of problems, for instance as variants of behavior cloning, or as a component of other algorithms such as DAGGER, policy distillation or KL-regularized RL. Our approach, \textit{augmented policy cloning} (APC), uses synthetic states to induce feedback-sensitivity in a region around sampled trajectories, thus dramatically reducing the environment interactions required for successful cloning of the expert. We achieve highly data-efficient transfer of behavior from an expert to a student policy for high-degrees-of-freedom control problems. We demonstrate the benefit of our method in the context of several existing and widely used algorithms that include policy cloning as a constituent part. Moreover, we highlight the benefits of our approach in two practically relevant settings (a) \emph{expert compression}, i.e. transfer to a student with fewer parameters; and (b) transfer from \emph{privileged experts}, i.e. where the expert has a different observation space than the student, usually including access to privileged information.
\end{abstract}
\section{Introduction}

\begin{figure*}
    \centering
    \includegraphics[width=0.9\textwidth]{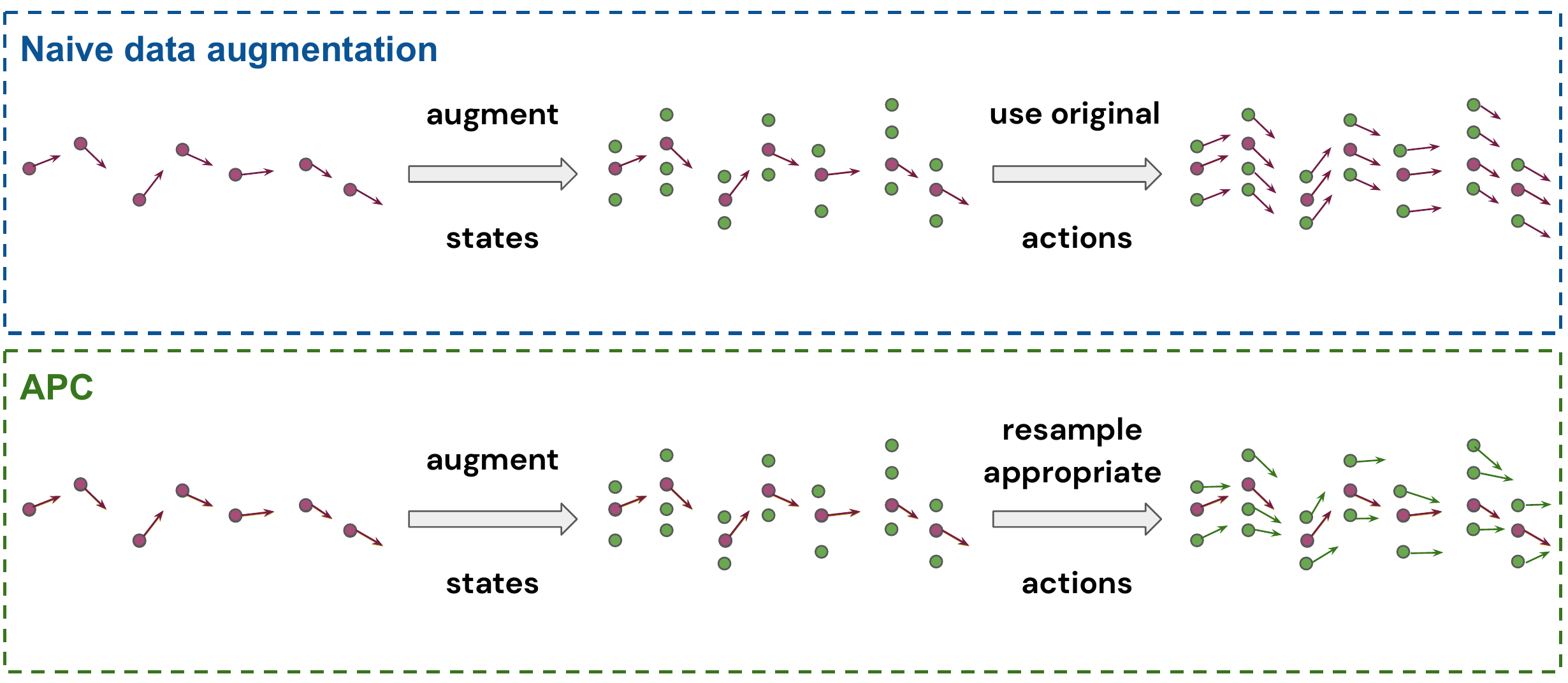}
    \caption{\textbf{Schematic of APC} (ours) method compared with a naive data augmentation approach. The original states (magenta circles) and actions (magenta arrows) pairs are then augmented by new virtual states (green circles). In Naive data augmentation, the same actions (magenta arrows) are used for all new virtual states. In APC, however, for each new virtual state, we resample a new action (green arrow) from the expert policy.} 
    \label{fig:schematic}
\end{figure*}

In various control and reinforcement learning settings, there is a need to transfer behavior from an expert policy to a student policy.  Broadly, when only samples from the expert policy are available, the standard approach is to employ a version of regression from states to actions. This class of approaches for producing a policy is known as behavioral cloning \citep{pomerleau1989alvinn, donald1996bc}.  Behavioral cloning is quite flexible and supports the setting where the expert trajectories come from a human teleoperating the relevant system directly, as well as various settings where the trajectories are sampled from other controllers, which themselves may have been trained or scripted.  However, for any of the settings where the expert policy is actually available, rather than just samples from the expert, it is reasonable to suspect that sampling random rollouts from the expert policy followed by performing behavioral cloning is not the most efficient approach for transferring behavior from the expert to the student.  Once a trajectory has been sampled via an expert rollout, there is actually additional information available that can be ascertained in the neighborhood of the trajectory, without having to perform an additional rollout, via the local feedback properties of the expert.

We refer to this setting, where we want to transfer from an expert policy to a student policy, while assuming the expert policy can be queried, as \textit{policy cloning}.  Naturally, there is still often an incentive to reduce the total number of rollouts, which may require actually collecting data in an unsafe or costly fashion, especially for real-world control problems. As such, there is a motivation to characterize any efficiency that can be gained in learning from small numbers of rollouts without as much concern for how many offline queries are required of the expert policy.  

If one has primarily encountered behavioral cloning in the context of learning from human demonstrations, policy cloning, with an available expert policy may seem contrived.  However, policy cloning naturally arises in many settings. For example, an expert policy may be too large to execute due to memory considerations as in \cite{parisotto2021efficient}, where authors propose to distill a large transformer network into small MLPs to be able to execute the policy on data collection workers. In a different setting, we could aim to distill an expert which has access to additional privileged information into a student without access to it, for example by distilling a full state policy into the one which has vision observations. And in the \textsc{DAgger} setting \citep{ross2011reduction}, a student policy collects data and is trained by regressing on the expert policy where state distribution comes from the student. 
In yet another setting, the expert may be suboptimal and the student needs to learn from expert while also being able to exceed the expert performance, perhaps by continuing to learn from a task via RL.  This problem has been described as \textit{kickstarting} in one incarnation \citep{schmitt2018kickstarting}, but also can arise when learning from behavioral priors \citep{tirumala2020behavior}, \citep{galashov2019information}, as also happens, for example, in Distral \citep{teh2017distral}. 
In all of the aforementioned situations, there is a motivation to minimize the amount of data a student needs to collect.

To improve data-efficiency in supervised settings generally, including in behavioral cloning settings, it is reasonable to consider data augmentation.  Data augmentation refers to applying perturbations to a finite training dataset to effectively amplify its diversity, usually in the hopes of producing a model that is invariant to the class of perturbations performed.  For example, in the well studied problem of object classification from single images, it is known that applying many kinds of perturbation should not affect the object label, so a model can be trained with many input perturbations all yielding the same output \citep{connor2019}.  This setting is fairly representative, with data augmentation usually intended to make the model ``robust" to nuisance perturbations of the input.  This class of image-perturbation has also been recently demonstrated to be effective in the context of control problems in the offline RL setting \citep{yarats2021image, laskin2020reinforcement}. 

Critically, for control problems it is not the case that the action should be invariant to the input state.  Or rather, while it does make sense for a control policy to be invariant to certain classes of sensor noise, an important class of robustness is that the policy is appropriately feedback-responsive.  This is to say that for small perturbations of the state of the control system, the optimal action is different in precisely the way that the expert implicitly knows.  This has been recognized and exploited in previous research that has distilled feedback-control plans into controllers \citep{mordatch2014combining, mordatch2015interactive, merel2019neural}. A similar intuition also underlies schemes which inject noise into the expert during rollouts to sample more comprehensively the space of how the expert recovers from perturbations \citep{laskey2017dart, merel2019neural}.  

In this work, we leverage this insight to develop a highly efficient policy cloning approach that makes use of both classes of data augmentation. For a high-DoF control problem that operates only from state (humanoid run and insert peg tasks from DeepMind control suite~\citep{tunyasuvunakool2020dm_control}), we demonstrate the feasibility of policy cloning that employs state-based data augmentation with expert querying to transfer the feedback-sensitive behavior of the expert in a region around a small number of rollouts.  Then on a more difficult high-DoF control problem that involves both state-derived and egocentric image observations (humanoid running through corridors task from DeepMind control~\citep{tunyasuvunakool2020dm_control}), we combine the state-based expert-aware data augmentation with a separate image augmentation intended to induce invariance to image perturbations.  Essentially our expert-aware data augmentation involves applying random perturbations to the state-derived observations, and training the student to match the expert-queried optimal action at each perturbed state,
thereby gaining considerable knowledge from the expert without performing excessive rollouts simply to cover the state space around existing trajectories. Our approach compares favorably to sensible baselines, including the naive approach of attempting to perform behavioral cloning with state perturbations, which
seeks to induce invariance (as proposed in \citealp{laskin2020reinforcement}) rather than feedback-sensitivity to state-derived observations. We demonstrate that our approach significantly improves data efficiency on all the settings mentioned above, i.e., behavioral cloning, \emph{expert compression}, cloning \emph{privileged experts}, \emph{DAgger} and \emph{kickstarting}.

\section{Problem description}
\label{sec:Problem}

\subsection{Expert-driven learning}

We start by introducing a notion of expert-driven learning that will be used throughout the paper. At first, we present a general form of the expert-driven objective and then introduce a few concrete examples. We consider a standard Reinforcement Learning (RL) problem. We present the domain as an MDP with continuous states for simplicity, however the problem definition is similar for a POMDP with observations derived from the state.
Formally, we describe the MDP in terms of a continuous state space $\mathcal{S} \in \mathcal{R}^{n}$ (for some $n>0$), an action space $\mathcal{A}$, transition dynamics $p(s' | s,a) : \mathcal{S} \times \mathcal{A} \rightarrow p(\mathcal{S})$, and a reward function $r : \mathcal{S} \times \mathcal{A} \rightarrow \mathcal{R}$. Let $\Pi$ be a set of parametric policies, i.e. of mappings $\pi_\theta : \mathcal{S} \rightarrow p(\mathcal{A})$ from the state space $\mathcal{S}$ to the probability distributions over actions $\mathcal{A}$, where $\theta \in \mathcal{R}^{m}$ for some $m>0$. For simplicity of the notation, we omit the parameter in front of the policy, i.e. $\pi = \pi_{\theta}$ and optimizing over the set of policies would be equivalent to the optimizing over a set of parameters. A reinforcement learning problem consists in finding such a policy $\pi$ that it maximizes the expected discounted future reward:
\begin{equation}
\label{eq:rl}
J(\pi) = \mathbb{E}_{p(\tau)}\left[\sum_{t} \gamma^{t} r(a_{t} | s_{t})\right],
\end{equation}
where $p(\tau) = p(s_{0}) \prod_{t}p(a_{t} | s_{t}) p(s_{t+1} | s_{t}, a_{t})$ is a trajectory distribution. We assume the existence of an expert policy $\pi_{E}(a | s)$.
This policy could be used to simplify the learning of a new policy on the same problem. Formally, we construct a new learning objective which aims to maximize the expected reward of the problem at hand as well as to clone the expert policy:
\begin{equation}
    \label{eq:expert_driven_rl}
    J(\pi, \pi_{E}) = \alpha J(\pi) - \lambda D(\pi, \pi_{E}),
\end{equation}
where $D$ is a function which measures the closeness of $\pi$ to $\pi_{E}$ and $\alpha \geq 0, \lambda \geq 0$ are parameters describing importance of both objectives. In most of the applications, $\alpha \in \{0, 1 \}$ and $\lambda \geq 0$ represents a relative importance of cloning an expert policy with respect to the RL objective.

\begin{figure*}[ht]
    \centering
    \includegraphics[width=0.93\textwidth]{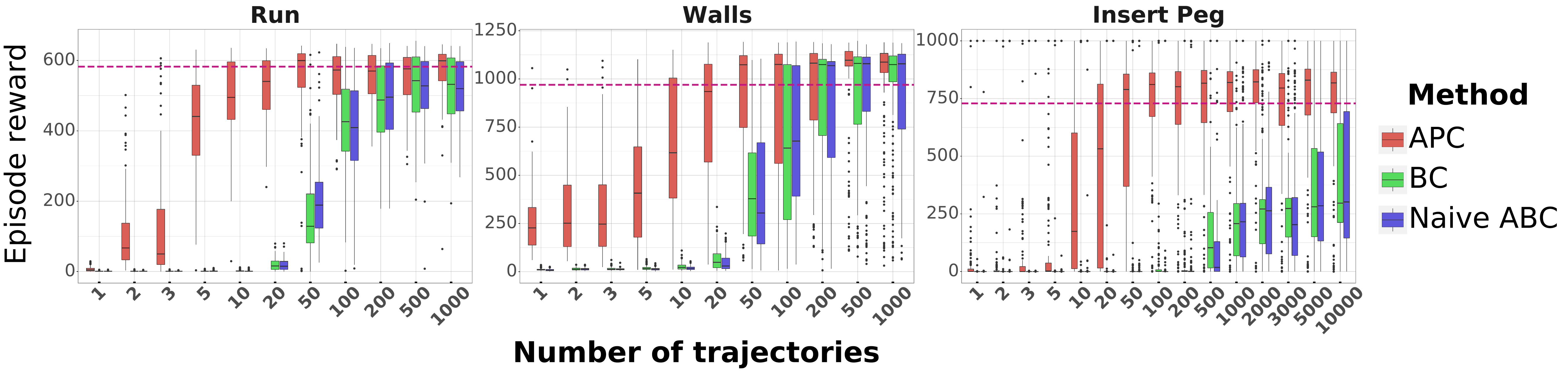}
    \caption{\textbf{Offline expert cloning results}. The X-axis represents the number of trajectories, the Y-axis corresponds to the episodic reward across 150 independent evaluations. The red color corresponds to APC, the blue to Naive ABC and the green to BC. The purple line depicts the average performance of the teacher policy.
    Each subplot represents a different task.} 
    \label{fig:bc_results}
\end{figure*}

\subsection{Behavioral cloning (BC)}
\label{sec:bc}
Behavioral cloning (BC) corresponds to optimizing the objective~\eqref{eq:expert_driven_rl} with $\alpha=0$, $\lambda=1$ and with $D$ defined as: 
\begin{equation}
    \label{eq:bc}
    D_{BC}(\pi, \pi_{E}) =  -\mathbb{E}_{(a,s) \in \mathcal{B}_{E}} [\log \pi(a|s) ]
\end{equation}
Here, $\mathcal{B}_{E}=\{(s_{i}, a_{i}), i=1,\ldots,N\}$, $N >0$ is a fixed dataset containing expert data. Minimizing the objective \eqref{eq:bc} is equivalent to maximizing the likelihood of the expert data under the policy $\pi$. The action $a$ in eqn.~\eqref{eq:bc} can be replaced by $\pi_{E}(s)$ for deterministic policies or by the mean $\mu_{E}(s)$ for Gaussian policies $\pi_{E}(\cdot | s) = \mathcal{N}(\mu_{E}(s), \sigma_{E}(s))$.

\subsection{Expert compression}
\label{sec:expert_compression}
In case of expert compression, we are interested in optimizing a similar objective as in eqn.~\eqref{eq:bc}, but where the student $\pi$ has smaller number of parameters compared to $\pi_{E}$.

\subsection{Learning from privileged experts}
\label{sec:privileged_expert}
In case of learning from privileged experts, we optimize similar objective to eqn.~\eqref{eq:bc}, where student receives different observations from the expert $\pi_{E}$. We assume that the expert has access to the privileged information, but the student does not. In particular, we consider the case where dataset with expert data contains observations (rather than full state), i.e. $\mathcal{B}_{E} =\{(o_{i}, a_{i}), i=1,\ldots,N\}$, $N >0$, but the expert has access to the full state $s_{i}$. More precisely, we consider an expert that observes the state $s=(s_{common}, s_{priv})$, where $s_{common}$ is a set of observations common to both the student and expert, whereas $s_{priv}$ is privileged information (containing some task-specific information). Then, the observations for the student are obtained as $o=(s_{common}, o_{vis})$ where $o_{vis}$ is the vision-based input.

\subsection{\textsc{DAgger}}
\label{sec:dagger}
Performance of Behavioral Cloning (BC) can be limited due to the fixed dataset, since the resulting policy may fail to generalize to states outside the training distribution. A different approach, known in the literature as \textsc{DAgger}~\citep{ross2011reduction} was proposed to overcome this limitation. In this setting, the expert is queried in states visited by the student, thus reducing distribution shift. In our notation, this corresponds to $\alpha=0$, $\lambda=1$ in eqn.~\eqref{eq:expert_driven_rl} and D is defined as:
\begin{equation}
    \label{eq:dagger}
    D_{\textsc{DAgger}}(\pi, \pi_{E}) = -\mathbb{E}_{p_{\beta}(\tau)} [\log \pi(a'_{t}|s_{t})],
\end{equation}
where $p_{\beta}(\tau)$, $\beta \in [0,1]$ is a trajectory distribution where actions are sampled according to the mixture policy between a student and an expert:
\begin{equation}
    \label{eq:dagger_mixture}
    p_{\beta}(a|s) = \beta \tilde{\pi}(a|s) + (1-\beta) \pi_{E}(a|s),
\end{equation}
The action $a'_{t}$ in eqn.~\eqref{eq:dagger} is obtained from the expert policy as $a'_{t} \sim \pi_{E}(\cdot | s_{t})$, as $\pi_{E}(s)$ for deterministic or as $\mu_{E}(s)$ for Gaussian policies $\pi_{E}(\cdot | s) = \mathcal{N}(\mu_{E}(s), \sigma_{E}(s))$ (see Section~\ref{sec:bc}). The policy $\tilde{\pi}(a|s)$ corresponds to a frozen version of student policy $\pi$ so that the gradient $\nabla_{\pi} D_{\textsc{DAgger}}(\pi, \pi_{E})$ ignores the acting distribution $p_{\beta}(a|s)$. Note that even though, in eqn.~\eqref{eq:dagger} we collect data from the environment, the setting nevertheless corresponds to pure imitation learning since  expected reward is not directly maximized.

\begin{algorithm}[ht]
\caption{Augmented Policy Cloning (APC)}
\label{alg:apc}
\begin{algorithmic}
    \STATE Parametric student policy: $\pi_{\theta}$
    \STATE Initial parameters: $\theta_{0}$
    \STATE Expert policy: $\pi_{E}$
    \STATE Dataset $\mathcal{B}_{E}=\{(s_{i}, a_{i}), i=1,\ldots,N\}$, $N >0$ of expert state-action pairs
    \STATE State perturbation noise $\sigma_{s}$
    \STATE Learning rate $\alpha$
    \STATE Number of augmented samples: $M$
    \STATE Number of gradient updates: $K$
    \STATE Size of a batch: $L$
    \FOR{k=1,\ldots,K}
    \STATE Sample a batch of pairs $\{(a_{i},s_{i})\}_{i=1}^{L} \sim \mathcal{B}_{E}$ 
    \STATE For each state $s_{i}$, sample $M$ perturbations $\delta s_{j} \sim \mathcal{N}(0, \sigma_{s}), j=1,\ldots,M$
    \STATE Construct $M$ virtual states $s'_{i, j} = s_{i} + \delta s_{j}, i=1,\ldots,L, j=1,\ldots,M$
    \STATE Resample new actions from expert $a'_{i,j} \sim \pi_{E}(\cdot | s'_{i, j})$
    \STATE For Gaussian experts, the action $a_{i} = \mu_{E}(s_{i})$ and the new actions are $a'_{i,j} = \mu_{E}(s'_{i,j})$
    \STATE Compute the empirical negative log-likelihood:
    \STATE \begin{center}
        $\mathcal{L} = - \left[ \log \pi_{\theta_{k}}(a_{i} | s_{i}) + \frac{1}{M}\sum_{j=1}^{M} \log \pi_{\theta_{k}}(a'_{i,j} | s'_{i, j}) \right]$
        \end{center}
    \STATE Update the parameters $\theta_{k+1} = \theta_{k} - \alpha \nabla_{\theta} \mathcal{L} $
    \ENDFOR
\end{algorithmic}
\end{algorithm}

\subsection{Kickstarting}

In eqn.~\eqref{eq:expert_driven_rl}, we combine both maximization of expected task reward and minimization of distance to the expert. In literature, it is known as \emph{Kickstarting}~\citep{schmitt2018kickstarting}. In this case, the objective from eqn.~\eqref{eq:expert_driven_rl} becomes:
\begin{equation}
    \label{eq:kickstarting}
    J(\pi, \pi_{E}) = J(\pi) - \lambda  \mathbb{E}_{p(\tau)} 
    \left[-\mathbb{E}_{\pi_{E}(a|s)} \log \pi(a|s)\right]
\end{equation}

where $p(\tau)$ is a trajectory distribution, where actions are sampled according to the student policy $\pi(\cdot|s)$. It corresponds to having $\alpha=1$, and $\lambda \geq 0$ and $D$ be state-conditional (across trajectory) cross-entropy from expert to a student.
Usually, in the \emph{Kickstarting} setting, the expert is sub-optimal and the goal is to train a policy that eventually outperforms the expert. Thus, it is customary to reduce $\lambda$ over the course of training. Yet, for simplicity, in our experiments we keep this coefficient fixed.

\begin{figure*}[ht]
     \centering
    \includegraphics[width=0.95\textwidth]{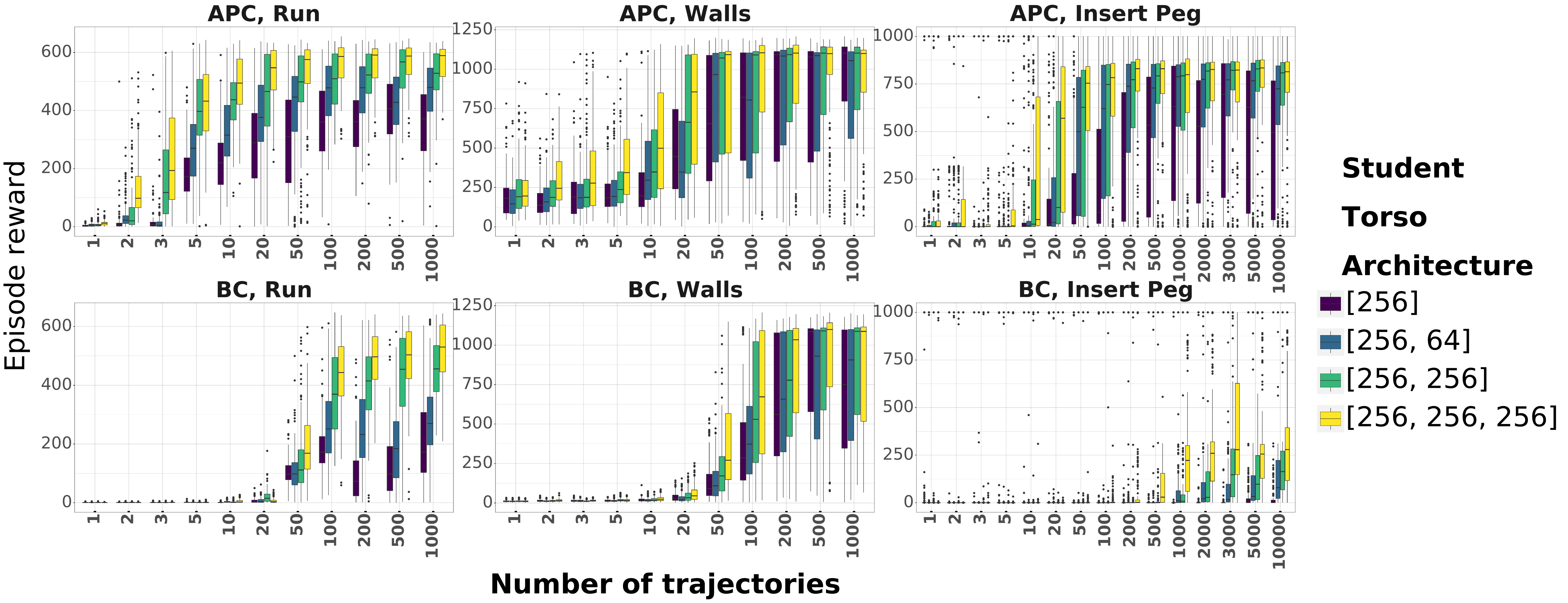}
    \caption{\textbf{\emph{Teacher compression} results}. We plot performances of different methods (rows) on different tasks (columns) as a function of number of trajectories available in the expert dataset. The legend corresponds to different student architectures used. We see that the performance degrades much more drastically for BC compared to APC. We report more complete results in Appendix~\ref{sec_app:small_network_add_results}}
    \label{fig:bc_small_networks_num_traj}
\end{figure*}

\section{Augmented policy cloning}
\label{sec:apc}

The previous section has demonstrated how the goal of cloning expert behavior can arise in different scenarios. In this section we propose a new and simple method which can significantly improve the data efficiency in the settings described in Section~\ref{sec:Problem}. We explain the basic idea for BC, but its generalization to other expert-driven learning approaches described in Section~\ref{sec:Problem} is straightforward. In Section~\ref{sec:additional_results} we show results for these problems.

When optimizing the objective \eqref{eq:bc}, for every state $s \in \mathcal{D}_{E}$ from the expert trajectories dataset, we consider a small Gaussian state perturbation:
\begin{equation}
    \label{eq:gaussian_perturbation}
    	\delta s \sim \mathcal{N}(0, \sigma^{2}_{s})
\end{equation}
which produces a new virtual state:
\begin{equation}
    \label{eq:state_perturbation}
    s' = s + \delta s 
\end{equation}
Then, for this state we query the expert and obtain a new action
\begin{equation}
    \label{eq:apc_action}
    a' \sim \pi_{E}(\cdot | s + \delta s)
\end{equation}
We then augment the dataset $\mathcal{B}_{E}$ with these new pairs of virtual states and actions. More explicitly the idea can be expressed in terms of the following objective:
\begin{multline}
    \label{eq:apc}
    D(\pi, \pi_{E})_{APC} = \mathbb{E}_{(a,s) \in \mathcal{B}_{E}} [\log \pi(a|s) + \\ \mathbb{E}_{\delta s \sim \mathcal{N}(0, \sigma^{2}_{s}), a' \sim \pi_{E}(\cdot | s + \delta s)}\log \pi(a' |s + \delta s)]
\end{multline}
We call this approach \textit{Augmented Policy Cloning} (APC) as it queries the expert policy to augment the training data. This approach is different from a naive data-augmentation technique, where a new state would be generated, but associated with the original action (and not a new one). It therefore allows to build policies which are feedback-responsive with respect to the expert. We illustrate it in Figure~\ref{fig:schematic} and we formulate APC algorithm for BC in Algorithm \ref{alg:apc}.

\section{Experimental details}

In this section we provide details common to all experiments. We provide additional details for each set of results at the beginning of Section~\ref{sec:results} and Section~\ref{sec:additional_results}.

\subsection{Domains}
\label{sec:domains}

To study how our method performs on complex control domains, we consider three complex, high-DoF continuous control tasks: \textit{Humanoid Run}, \textit{Humanoid Walls} and \textit{Insert Peg}. All these domains are implemented using the MuJoCo physics engine \citep{todorov2012mujoco} and are available in the \texttt{dm\_control} repository \citep{tunyasuvunakool2020dm_control}. These problems are rather challenging, requiring stabilization of a complex body (for humanoid tasks), vision to guide the movement (\textit{Walls} task), and solving a complex control problem with a weak reward signal (\textit{Insert Peg}). These environments are related to the domains that have been proposed for use in offline RL benchmarks \citep{gulcehre2020rl}; however, the experiments we perform in this work require availability of the expert policy, so we do not use offline data, but instead train new experts and perform experiments in the very low data regime. We compare all methods on \textit{Humanoid Run} and \textit{Humanoid Run} tasks and report report a subset of results on \textit{Insert Peg}, due to complex nature of the experiments. For more details, please refer to Appendix~\ref{sec_app:environments}.

\begin{figure*}[ht]
     \centering
    \includegraphics[width=0.85\textwidth]{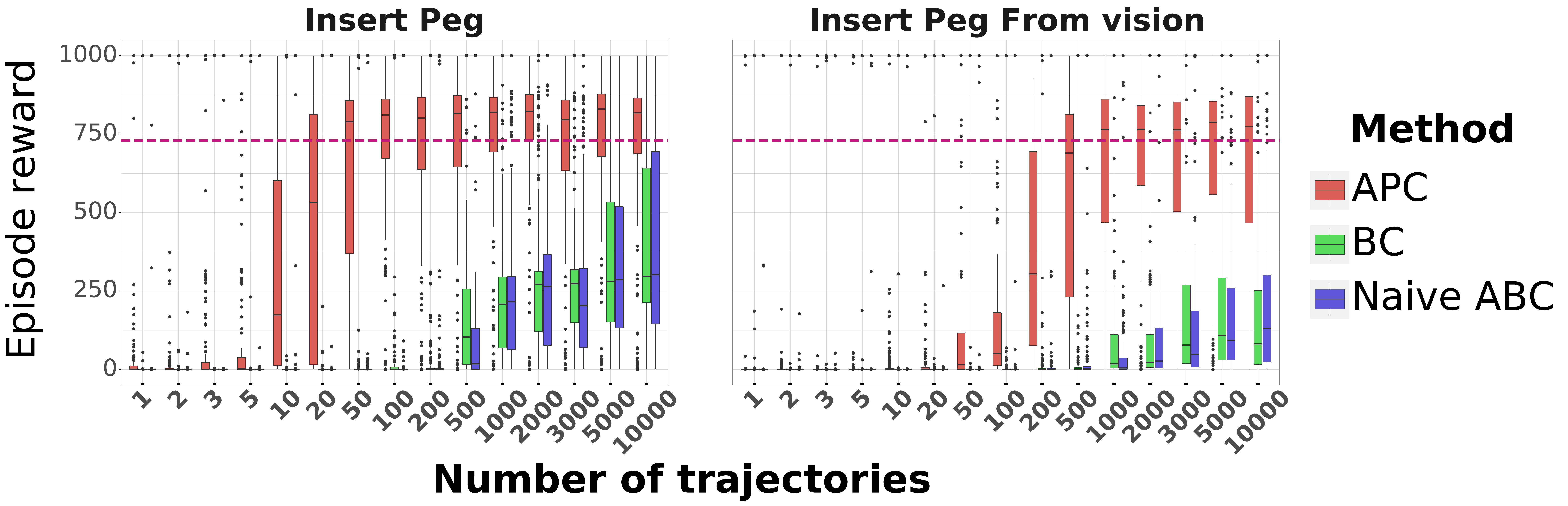}
    \caption{\textbf{Learning from privileged experts results}. On the left plot we present the performance of different methods on \textit{Insert Peg} task where additional privileged information (target position) is available. On the right, we present the results where instead of privileged information (target position), we use visual input. We observe that learning from vision is less data efficient and more complicated, but APC manages to obtain comparable performance to the scenario with privileged information, whereas BC and Naive ABC fail to learn.}
    \label{fig:state_to_vision_results}
\end{figure*}

\subsection{Baselines}
\label{sec:baselines}

As baselines we consider simple BC as described in eqn.~\eqref{eq:bc} as well as a simple modification of BC, where, similarly to APC, we apply state perturbations to expert trajectories as in eqn.~\eqref{eq:gaussian_perturbation} and eqn.~\eqref{eq:state_perturbation}, but we do not produce a new action from the expert (i.e., we augment the states but keep the same action). We call this approach Naive Augmented Behavioral Cloning (Naive ABC). Essentially, this method trains a student policy to produce the same action in response to small state perturbations.  This approach is motivated by analogy to how one might build robustness in a classifier. However, this is naive when applied to continuous control problems where even small changes in input should lead to a change in action.  Moreover, for \textit{Humanoid Walls} task, we considered additional random crops augmentations applied to visual input of the student (not the expert) which is similar in spirit to \cite{laskin2020reinforcement}. Note that in this case, it would also robustify the student to these vision augmentations as it will not produce a new action even in the APC (since the expert was not trained with data augmentations). When vision augmentations are used together with either APC or naive ABC, we add "with image" to the method name. When only image augmentations are used (without any state-based augmentations), we call it "image only". The purpose of combining vision and state augmentations is to study the interplay between APC and more traditional data augmentation methods. We only report results with vision augmentations for \textit{DAgger} and \textit{kickstarting} in the main paper and we provide additional offline policy cloning results in Appendix~\ref{sec_app:walls_additional_results}.

\section{Core results: offline policy cloning}
\label{sec:results}

\begin{figure*}[ht]
    \centering
    \includegraphics[width=0.8\textwidth]{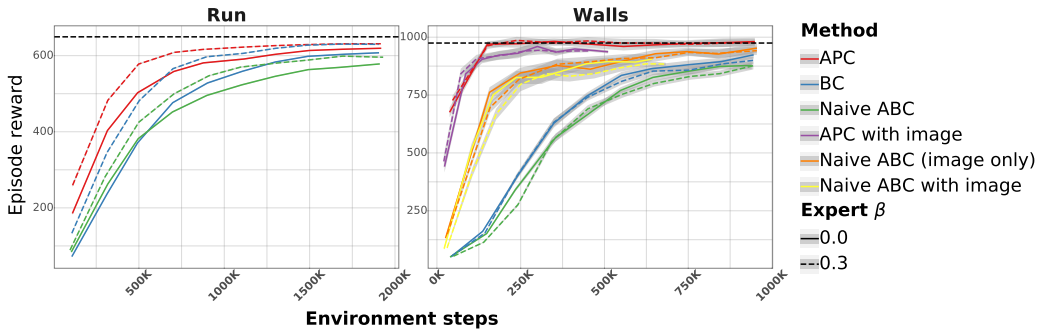}
    \caption{\textbf{\textsc{DAgger} results}. On the X-axis we report the number of environment steps. On the Y-axis we report averaged across 3 seeds episodic reward achieved by the student. We report confidence intervals in the shaded areas. For Run task, the confidence intervals are very small and are not visible.
    In solid line we report the performance without using expert policy during the acting. In dashed line, we report the performance of the policy which mixes 30\% with the expert. All the methods use mean action during evaluation.
    }
    \label{fig:dagger_results}
\end{figure*}

\paragraph{Training and evaluation protocols.}
We train expert policies till convergence using MPO algorithm~\cite{abdolmaleki2018maximum} for \textit{Humanoid} tasks and VMPO algorithm~\cite{song2019vmpo} for \textit{Insert Peg} task, as we found MPO was unable to learn on this task. The policies are represented by the Gaussian distribution $\pi_{E}(\cdot | s) = \mathcal{N}(\mu_{E}(s), \sigma(s))$.
We create datasets as in eqn.~\eqref{eq:bc} using pre-trained experts with a different number of expert trajectories. To asses the sensitivity of different methods to the expert noise, when constructing a dataset, the expert action is drawn according to Gaussian distribution with a fixed variance, i.e. $a \sim \mathcal{N}(\mu_{E}(s), \sigma_{E})$, 
where $\sigma_{E}$ is the fixed amount of expert noise. In the subsequent BC experiments, we use $\sigma_{E}=0.2$. Moreover, in order to analyze the noise robustness of the student policy is trained via BC, $\pi(\cdot|s) = \mathcal{N}(\mu(s),\sigma(s))$, we evaluate it by executing the action drawn from a Gaussian with a fixed variance, i.e. $a \sim \mathcal{N}(\mu(s), \sigma)$, where $\sigma$ is the fixed amount of student noise. In all the experiments below we use $\sigma=0.2$. We apply early stopping and select hyperparameters based on the evaluation performance on a validation set. We always report performance based on 150 random environment instantiations. For more details, see Appendix~\ref{sec_ap:offline_policy_cloning}.

\subsection{Applying Augmented Policy Cloning}
We evaluate the performance of APC when fitting the fixed dataset of expert trajectories. For the APC method, we rely on Algorithm~\ref{alg:apc}. We use baselines described in Section~\ref{sec:baselines}. In Figure~\ref{fig:bc_results}, we show the performance of different methods on different tasks as a function of number of trajectories available in the dataset. We see that APC requires significantly fewer expert trajectories to achieve a high level of performance. 
Moreover, we see that Naive ABC performs very similarly to BC. 
The results suggest that when cloning an expert using a small fixed set of states APC can provide significant advantages.

In Appendix~\ref{sec_app:short_traj_experiment}, we report additional results for a scenario, where instead of full long trajectories for each task, we consider only short trajectories (i.e., the early portion of episodes). The motivation for this experiment is to see whether we can further improve the data efficiency of the methods. Moreover, in domains where the initial state distribution is randomized in meaningful ways, shorter trajectories can provide an advantage because we get to observe more diverse initial states. Incidentally, this supplemental comparison shows that for \textit{Insert Peg} all methods performed better with short trajectories, because initial snippets of episodes actually include the full solution to the task (i.e., the expert policy rapidly inserts the peg and the episode doesn't immediately terminate).  We report only long trajectories in Figure~\ref{fig:bc_results}, since APC performs relatively well in both cases and full length trajectories correspond to the most straightforward setting.

\subsection{Expert compression}
To study APC in a practically motivated setting we consider  \emph{expert compression} as discussed in Section~\ref{sec:expert_compression}, where a student policy has fewer parameters than the expert. This setting occurs, for instance,  when the system is subject to computational contstraints (time, memory, etc.), as in \citep{parisotto2021efficient}. To study APC's data efficiency in this setting, we consider different sizes of the student network torso, where $[256, 256, 256]$ corresponds to the original network size (see Appendix~\ref{sec_app:expert_compression_details} for more details).

The results are given in Figure~\ref{fig:bc_small_networks_num_traj}. We observe that the performance of all methods degrades when the student network is smaller than the original one, but the degradation is much less severe for APC, while maintaining high level of data efficiency. See Appendix~\ref{sec_app:small_network_add_results} for more complete results (with naive ABC) as well as additional ablation over student network torso sizes.

\subsection{Learning from privileged experts.}
Next, we consider a scenario where the expert has access to \emph{privileged information} that is not available to the student, as discussed in Section~\ref{sec:privileged_expert}.
To study the impact of APC in this scenario, we train the expert on \textit{Insert Peg} task where the full state contains common information (proprioception, sword position and orientation) and privileged information of the target position. The student is given access to the common observations as well as a third person (camera) view of the scene providing information about the target position.
The latter setup is similar in spirit to \citep{laskin2020reinforcement} For more details, see Appendix~\ref{sec_app:priveledged_details}.

The results are given in Figure~\ref{fig:state_to_vision_results}. We observe that APC achieves similar performance in both settings, provided a sufficient amount of trajectories are available, whereas BC and Naive ABC fail to transfer the expert's behavior to the student.

\section{Additional Results: Augmented Policy Cloning as a subroutine}
\label{sec:additional_results}

\begin{figure*}[ht]
    \centering
    \includegraphics[width=0.8\textwidth]{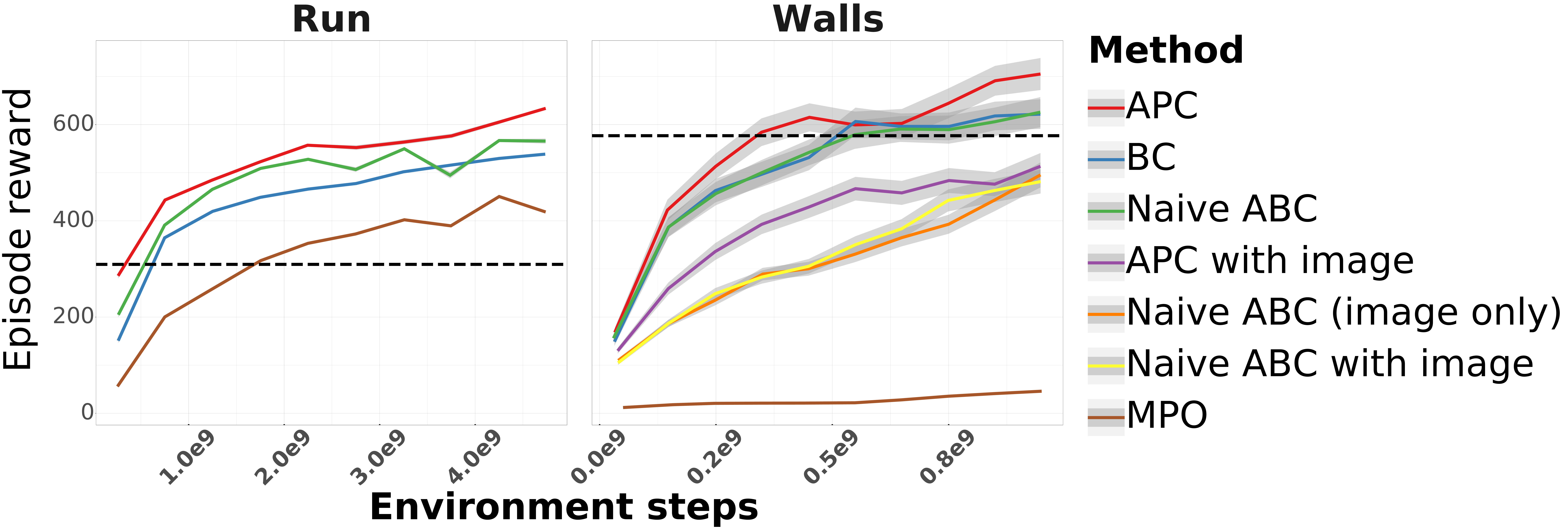}
    \caption{\textbf{Kickstarting results}. On the X-axis we show the number of environment steps and on the Y-axis we report averaged across 3 seeds episodic reward achieved by the student. We report confidence intervals in the shaded areas. For Run task, these intervals are small and are not visible. Dashed black line shows the expert performance.} 
    \label{fig:kickstarting_results}    
\end{figure*}

\subsection{\textsc{DAgger} with data augmentation}
\label{sec:dagger_results}
As described in Section~\ref{sec:dagger}, \textsc{DAgger}~\citep{ross2011reduction} is a more sophisticated approach where data is collected from the real environment by executing a policy from eqn.~\eqref{eq:dagger_mixture}, which is a mixture between a student and an expert. In this section we study how data augmentation approaches affect the data efficiency of the \textsc{DAgger} algorithm.

For each task, we train expert policies to convergence using the MPO algorithm~\cite{abdolmaleki2018maximum}. We consider similar baselines for both tasks as in the previous section. For an expert policy that has been pre-trained via MPO~\citep{abdolmaleki2018maximum}, we perform online rollouts for two values of the expert-student mixing coefficient, $\beta =0$ and $\beta=0.3$ (see eqn.~\ref{eq:dagger_mixture}).
Since both student and expert are Gaussian distributions, instead of using a $\log \pi$ in eqn.~\eqref{eq:dagger}, we could use a state-conditional cross entropy from an expert to a student, $\mathcal{H}[\pi_{E}(\cdot | s) || \pi(\cdot | s)]$. Empirically, we found that it worked better than using $\log \pi$ (see Appendix~\ref{sec:qual_comp}). We run experiments in a data-restricted setup. For more details, see Appendix~\ref{sec_app:dagger_details}.

Results are shown in Figure~\ref{fig:dagger_results}. We see that APC and its vision variant outperform BC and Naive ABC similarly to the behavior cloning experiments. 
While we observe that image augmentation can help, we see that the primary advantage comes from the state-based augmentation for APC. For the Run task, we observe that all \textsc{DAgger} methods achieve slightly lower performance than an expert policy. We speculate that this is due to insufficient coverage of the state space during training.

\subsection{Kickstarting with data augmentation.}
\label{sec:kickstarting_results}
A similar in spirit approach is kickstarting~\cite{schmitt2018kickstarting}, where we solve an RL task as well as cloning the expert policy.
Similarly to previous section, we apply APC in kickstarting on the cross entropy term in eqn.~\eqref{eq:kickstarting}.

For each task, we train expert policies to convergence using the MPO algorithm~\cite{abdolmaleki2018maximum}. Since in the kickstarting we are interested in outperforming a sub-optimal expert, for each task, we select experts such that they achieve around 50 \% of the optimal performance. On top of kicktarting, we report the performance of MPO~\cite{abdolmaleki2018maximum} learning from scratch on the task of interest. We run experiment in a distributed, high data regime. All the details are given in Apendix~\ref{sec_app:kickstarting_details}.

The results are given in Figure~\ref{fig:kickstarting_results}. 
We observe that APC performs better than Naive ABC on \textit{Humanoid Run} task and similarly on \textit{Humanoid Walls} task. Both approaches perform better than BC and learning from scratch. We hypothesise that the reason of not seeing a consistent advantage could be due several factors. Firstly, as we are in a high-data and distributed regime, since there is no limit on relative acting / learning ratio, and acting policies are not restricted to collect trajectories, it is unclear whether data-augmentation should help. We tried to explore the rate-limiting regime, but we experienced instability of kickstarting experiments. Secondly, we use reward signal which makes the impact of expert cloning less important. Thirdly, on top of learning the policy, we also need to learn an state-action value $Q(s,a)$ function. Unfortunately, we cannot use APC-style data augmentation for learning $Q$ function, therefore it might be the bottleneck.
Finally, unlike in kickstarting~\cite{schmitt2018kickstarting}, we do not use an annealing schedule of $\lambda$ to make the experiments simpler, but we still observe that a fixed coefficient helps to kickstart an experiment and outperform an expert policy. On top of that, we see that image-based augmentation have less of impact in this setting and generally leads to poor performance.

\section{Discussion}
\label{sec:discussion}

Many expert-driven learning approaches actually have access to an expert that can be queried; however, this opportunity is rarely exploited fully. In this work we demonstrated a general scheme for more efficient transfer of expert behavior by augmenting expert trajectory data with virtual, perturbed states as well as the expert actions in these virtual states. This data augmentation technique is widely applicable and we demonstrated that it improves data efficiency when used in place of behavioral cloning in various settings including offline cloning, expert compression, transfer from privileged experts, or when behavioral cloning is used as a subroutine within online algorithms such as \textsc{DAgger} or kickstarting.

Critically, data efficiency is generally very important in realistic applications, where new data acquisition cost could be high. In particular, settings involving deployment of policies in the real world, such as robotics applications, may benefit from an ability to efficiently transfer expert policy behavior from one neural network to another (for compression or execution speed reasons). While overall, we consider the present work to be fairly basic research with limited ethical impact, insofar as our approach decreases the amount of data which needs to be collected through processes which could potentially be unsafe or costly, there is a potential positive social value.

Our approach is neither intended for nor suitable for all control settings.  Fundamentally, our approach relies upon the ability to query expert policy for the perturbed states.  This arises frequently enough to be worth our investigation, but is a limiting assumption.  Our approach was also developed with continuous control problems, essentially with continuous observation spaces as well as continuous action spaces in mind.  Related approaches may be worth pursuing in discrete control problems, but that has not been a focus of the present work.

An interesting future direction would be to explore different ways we could generate virtual states. Even though simple Gaussian perturbation of states seems to work fairly well, we could explore a possibility of building a state model and sample the virtual states from it.

\bibliography{main}

\begin{thebibliography}{21}
\providecommand{\natexlab}[1]{#1}
\providecommand{\url}[1]{\texttt{#1}}
\expandafter\ifx\csname urlstyle\endcsname\relax
  \providecommand{\doi}[1]{doi: #1}\else
  \providecommand{\doi}{doi: \begingroup \urlstyle{rm}\Url}\fi

\bibitem[Abdolmaleki et~al.(2018)Abdolmaleki, Springenberg, Tassa, Munos,
  Heess, and Riedmiller]{abdolmaleki2018maximum}
Abbas Abdolmaleki, Jost~Tobias Springenberg, Yuval Tassa, Remi Munos, Nicolas
  Heess, and Martin Riedmiller.
\newblock Maximum a posteriori policy optimisation, 2018.

\bibitem[Galashov et~al.(2019)Galashov, Jayakumar, Hasenclever, Tirumala,
  Schwarz, Desjardins, Czarnecki, Teh, Pascanu, and
  Heess]{galashov2019information}
Alexandre Galashov, Siddhant~M. Jayakumar, Leonard Hasenclever, Dhruva
  Tirumala, Jonathan Schwarz, Guillaume Desjardins, Wojciech~M. Czarnecki,
  Yee~Whye Teh, Razvan Pascanu, and Nicolas Heess.
\newblock Information asymmetry in kl-regularized rl, 2019.

\bibitem[Gulcehre et~al.(2020)Gulcehre, Wang, Novikov, Paine, G{\'o}mez, Zolna,
  Agarwal, Merel, Mankowitz, Paduraru, et~al.]{gulcehre2020rl}
Caglar Gulcehre, Ziyu Wang, Alexander Novikov, Thomas Paine, Sergio G{\'o}mez,
  Konrad Zolna, Rishabh Agarwal, Josh~S Merel, Daniel~J Mankowitz, Cosmin
  Paduraru, et~al.
\newblock Rl unplugged: A collection of benchmarks for offline reinforcement
  learning.
\newblock \emph{Advances in Neural Information Processing Systems}, 33, 2020.

\bibitem[Hoffman et~al.(2020)Hoffman, Shahriari, Aslanides, Barth-Maron,
  Behbahani, Norman, Abdolmaleki, Cassirer, Yang, Baumli, Henderson, Novikov,
  Colmenarejo, Cabi, Gulcehre, Paine, Cowie, Wang, Piot, and
  de~Freitas]{hoffman2020acme}
Matt Hoffman, Bobak Shahriari, John Aslanides, Gabriel Barth-Maron, Feryal
  Behbahani, Tamara Norman, Abbas Abdolmaleki, Albin Cassirer, Fan Yang, Kate
  Baumli, Sarah Henderson, Alex Novikov, Sergio~Gómez Colmenarejo, Serkan
  Cabi, Caglar Gulcehre, Tom~Le Paine, Andrew Cowie, Ziyu Wang, Bilal Piot, and
  Nando de~Freitas.
\newblock Acme: A research framework for distributed reinforcement learning,
  2020.

\bibitem[Laskey et~al.(2017)Laskey, Lee, Fox, Dragan, and
  Goldberg]{laskey2017dart}
Michael Laskey, Jonathan Lee, Roy Fox, Anca Dragan, and Ken Goldberg.
\newblock Dart: Noise injection for robust imitation learning.
\newblock In \emph{Conference on robot learning}, pages 143--156. PMLR, 2017.

\bibitem[Laskin et~al.(2020)Laskin, Lee, Stooke, Pinto, Abbeel, and
  Srinivas]{laskin2020reinforcement}
Michael Laskin, Kimin Lee, Adam Stooke, Lerrel Pinto, Pieter Abbeel, and
  Aravind Srinivas.
\newblock Reinforcement learning with augmented data, 2020.

\bibitem[Merel et~al.(2019)Merel, Hasenclever, Galashov, Ahuja, Pham, Wayne,
  Teh, and Heess]{merel2019neural}
Josh Merel, Leonard Hasenclever, Alexandre Galashov, Arun Ahuja, Vu~Pham, Greg
  Wayne, Yee~Whye Teh, and Nicolas Heess.
\newblock Neural probabilistic motor primitives for humanoid control, 2019.

\bibitem[Michie and Sammut(1996)]{donald1996bc}
Donald Michie and Claude Sammut.
\newblock \emph{Behavioural Clones and Cognitive Skill Models}, page 387–395.
\newblock Oxford University Press, Inc., USA, 1996.
\newblock ISBN 019853860X.

\bibitem[Mordatch and Todorov(2014)]{mordatch2014combining}
Igor Mordatch and Emo Todorov.
\newblock Combining the benefits of function approximation and trajectory
  optimization.
\newblock In \emph{Robotics: Science and Systems}, volume~4, 2014.

\bibitem[Mordatch et~al.(2015)Mordatch, Lowrey, Andrew, Popovic, and
  Todorov]{mordatch2015interactive}
Igor Mordatch, Kendall Lowrey, Galen Andrew, Zoran Popovic, and Emanuel~V
  Todorov.
\newblock Interactive control of diverse complex characters with neural
  networks.
\newblock \emph{Advances in Neural Information Processing Systems},
  28:\penalty0 3132--3140, 2015.

\bibitem[Parisotto and Salakhutdinov(2021)]{parisotto2021efficient}
Emilio Parisotto and Ruslan Salakhutdinov.
\newblock Efficient transformers in reinforcement learning using actor-learner
  distillation, 2021.

\bibitem[Pomerleau(1989)]{pomerleau1989alvinn}
Dean~A Pomerleau.
\newblock Alvinn: An autonomous land vehicle in a neural network.
\newblock Technical report, Carnegie-Mellon, 1989.

\bibitem[Ross et~al.(2011)Ross, Gordon, and Bagnell]{ross2011reduction}
Stephane Ross, Geoffrey~J. Gordon, and J.~Andrew Bagnell.
\newblock A reduction of imitation learning and structured prediction to
  no-regret online learning, 2011.

\bibitem[Schmitt et~al.(2018)Schmitt, Hudson, Zidek, Osindero, Doersch,
  Czarnecki, Leibo, Kuttler, Zisserman, Simonyan, and
  Eslami]{schmitt2018kickstarting}
Simon Schmitt, Jonathan~J. Hudson, Augustin Zidek, Simon Osindero, Carl
  Doersch, Wojciech~M. Czarnecki, Joel~Z. Leibo, Heinrich Kuttler, Andrew
  Zisserman, Karen Simonyan, and S.~M.~Ali Eslami.
\newblock Kickstarting deep reinforcement learning, 2018.

\bibitem[Shorten and Khoshgoftaar(2019)]{connor2019}
Connor Shorten and Taghi Khoshgoftaar.
\newblock A survey on image data augmentation for deep learning.
\newblock \emph{Journal of Big Data}, 6, 07 2019.
\newblock \doi{10.1186/s40537-019-0197-0}.

\bibitem[Song et~al.(2019)Song, Abdolmaleki, Springenberg, Clark, Soyer, Rae,
  Noury, Ahuja, Liu, Tirumala, Heess, Belov, Riedmiller, and
  Botvinick]{song2019vmpo}
H.~Francis Song, Abbas Abdolmaleki, Jost~Tobias Springenberg, Aidan Clark,
  Hubert Soyer, Jack~W. Rae, Seb Noury, Arun Ahuja, Siqi Liu, Dhruva Tirumala,
  Nicolas Heess, Dan Belov, Martin Riedmiller, and Matthew~M. Botvinick.
\newblock V-mpo: On-policy maximum a posteriori policy optimization for
  discrete and continuous control, 2019.

\bibitem[Teh et~al.(2017)Teh, Bapst, Czarnecki, Quan, Kirkpatrick, Hadsell,
  Heess, and Pascanu]{teh2017distral}
Yee~Whye Teh, Victor Bapst, Wojciech~Marian Czarnecki, John Quan, James
  Kirkpatrick, Raia Hadsell, Nicolas Heess, and Razvan Pascanu.
\newblock Distral: Robust multitask reinforcement learning.
\newblock \emph{arXiv preprint arXiv:1707.04175}, 2017.

\bibitem[Tirumala et~al.(2020)Tirumala, Galashov, Noh, Hasenclever, Pascanu,
  Schwarz, Desjardins, Czarnecki, Ahuja, Teh, et~al.]{tirumala2020behavior}
Dhruva Tirumala, Alexandre Galashov, Hyeonwoo Noh, Leonard Hasenclever, Razvan
  Pascanu, Jonathan Schwarz, Guillaume Desjardins, Wojciech~Marian Czarnecki,
  Arun Ahuja, Yee~Whye Teh, et~al.
\newblock Behavior priors for efficient reinforcement learning.
\newblock \emph{arXiv preprint arXiv:2010.14274}, 2020.

\bibitem[Todorov et~al.(2012)Todorov, Erez, and Tassa]{todorov2012mujoco}
Emanuel Todorov, Tom Erez, and Yuval Tassa.
\newblock Mujoco: A physics engine for model-based control.
\newblock In \emph{2012 IEEE/RSJ International Conference on Intelligent Robots
  and Systems}, pages 5026--5033. IEEE, 2012.

\bibitem[Tunyasuvunakool et~al.(2020)Tunyasuvunakool, Muldal, Doron, Liu,
  Bohez, Merel, Erez, Lillicrap, Heess, and
  Tassa]{tunyasuvunakool2020dm_control}
Saran Tunyasuvunakool, Alistair Muldal, Yotam Doron, Siqi Liu, Steven Bohez,
  Josh Merel, Tom Erez, Timothy Lillicrap, Nicolas Heess, and Yuval Tassa.
\newblock dm\_control: Software and tasks for continuous control.
\newblock \emph{Software Impacts}, 6:\penalty0 100022, 2020.

\bibitem[Yarats et~al.(2021)Yarats, Kostrikov, and Fergus]{yarats2021image}
Denis Yarats, Ilya Kostrikov, and Rob Fergus.
\newblock Image augmentation is all you need: Regularizing deep reinforcement
  learning from pixels.
\newblock In \emph{9th International Conference on Learning Representations,
  ICLR}, volume 2021, 2021.

\end{thebibliography}

\newpage
\appendix
\onecolumn

\section{Environment details}
\label{sec_app:environments}
In this work we consider three environments from DeepMind Control repository~\citep{tunyasuvunakool2020dm_control}: \textit{Humanoid Run}, \textit{Humanoid Walls} and \textit{Insert Peg}. \textit{Humanoid Run} task requires an agent controlling humanoid body to run at a specific speed and gets reward which is proportional to the inverse distance between its current speed and the target speed. The observations are based on proprioceptive information.
In \textit{Humanoid Walls}, an agent controls a humanoid body to run along a corridor and avoid walls, at highest possible speed. The observations are based on proprioception and on egocentric vision.  It receives reward which is proportional to the forward speed (through the corridor), thus incentivising it to run as fast as possible. It receives proprioceptive observations as well as the image of size 64x64 from the ego-centric camera. In our experiments, we use \textit{Simple Humanoid} body rather than \textit{CMU Humanoid} as in the original task in order to simplify the experiments. Action dimension is equal to 21. Finally, in \textit{Insert Peg}, an agent controls an arm which needs to put a sword into a narrow hole. The observations are based on proprioception, sword position and orientation, hole position. In case of vision-based input, we add second person camera observations with image size of 64x64. For more details, check \cite{tunyasuvunakool2020dm_control}. Note these environments are related to the domains that have been proposed for use in offline RL benchmarks \citep{gulcehre2020rl}; however, the experiments we perform in this work require availability of the expert policy, so we do not use offline data, but instead train new experts and perform experiments in the very low data regime. We choose these exact environments due to their challenging nature in order to demonstrate the impact of APC on data efficiency, but we did try initially simpler environments from the DeepMind Control repository~\citep{tunyasuvunakool2020dm_control} which included \textit{Humanoid Walk} and \textit{Walker Walk/Run} tasks. We decided not to conduct exhaustive experiments on these environments.

\section{Methods and baselines}
\label{sec_app:baselines}

The main method we consider is APC described in Section~\ref{sec:apc} for offline experts cloning experiments. For all the methods in offline expert cloning experiments, as an action in the objective from eqn.~\eqref{eq:bc}, we use an expert mean $\mu_{E}(s)$. We also extend this method to scenarios from Section~\ref{sec:additional_results}, i.e. on \textsc{DAgger} and kickstarting. For both scenarios, applying APC is straightforward. In case of \textsc{DAgger}, the APC approach would correspond to resampling additional states via eqn.~\eqref{eq:gaussian_perturbation} and via eqn.~\eqref{eq:state_perturbation} for each state $s_{t}$ encountered in the objective from the eqn.~\eqref{eq:dagger}. Then, for each such new state $s'=s_{t} + \delta s$, we would add a term to optimize in the objective, corresponds to the cross entropy form the expert to the student, i.e. $\mathcal{H}[\pi_{E}(\cdot | s') || \pi(\cdot | s')]$. We can also consider resampling a new action, but we empirically found that cross-entropy worked better, see Appendix~\eqref{sec:qual_comp}. For kickstarting, applying APC would also correspond to sampling new virtual states $s' = s + \delta s$ for each state in the second term of the eqn.~\eqref{eq:kickstarting}. Then similarly, we would add an additional cross entropy term to the objective, i.e., $\mathcal{H}[\pi_{E}(\cdot | s') || \pi(\cdot | s')]$.

As baselines against APC, we consider BC algorithm described in eqn.~\eqref{eq:bc}, which in \textsc{DAgger} and kickstarting simply corresponds to the unmodified versions of this method. On top of BC, we consider a simple modification of BC, where we apply, similar to APC, state perturbations to expert trajectories as in eqn.~\eqref{eq:gaussian_perturbation} and eqn.~\eqref{eq:state_perturbation}, but we do not produce a new action from the expert and use the original one. We call this approach Naive Augmented Behavioral Cloning (Naive ABC). Essentially, this method trains a student policy to be robust with respect to small state perturbations. The application of Naive ABC in case of \textsc{DAgger} and kickstarting is similar to APC, with the exception that we now consider the cross entropy term $\mathcal{H}[\pi_{E}(\cdot | s) || \pi(\cdot | s')]$, where the student is taken on the new augmented states and the expert on the original, unmodified ones.

Moreover, for \textit{Humanoid Walls} task, we considered additional vision-based augmentations, random crops, similar in spirit to \cite{laskin2020reinforcement}. Note that in this case, it would also robustify the student to these vision augmentations as it will not produce a new action even in the APC (since the expert was not trained with data augmentations). When vision augmentations are used together with APC or naive ABC, we add "with image" to the method name.  When only image augmentations are used (without any state-based augmentations), we call it "image only". The purpose of combining vision and state augmentations is to study the interplay between APC and more traditional data augmentation methods. We use random crops producing images of size 48x48 instead of the original 64x64 images.

\section{Agent architecture}
\label{sec:agent_architecture}

For all the experiments, we use the same agent architecture. The agent has two separate networks: actor (policy) and critic (Q-function). Both networks are split into 3 components: encoder, torso and head. Encoders for actor and critic are separate but have the same architecture. For state-only (no vision) observations, encoder corresponds to a simple concatenations of all the observations. For the visual input, it divides each pixel by $255$ and then applies a 3-layer ResNet of sizes $(16, 32, 32)$ with $ELU$ activations followed by a linear layer of size 256 and $ELU$ activation. The resulting output is then concatenated together with state-based input. For actor network, torso corresponds to a 3 dimensional MLP, each hidden layer of size $256$ with activation $ELU$ applied at the end of each hidden layer. The output of actor torso is then passed to the actor head network, which applies a linear layer (without activation) with output size equal to $N_{a} * 2$, where $N_{a}$ is the action dimension. It produces the actor mean $\mu$ and log-variance: $\log \tilde{\sigma}$. Then, the variance of the actor is calculated as
\begin{equation*}
    \sigma = soft plus (\log \tilde{\sigma})  + \sigma_{min},
\end{equation*}
where $\sigma_{min} = 0.0001$. That would encode the Gaussian policy $\pi(\cdot | s) = \mathcal{N}(\mu(s) | \sigma(s))$. This parameterization ensures that the variance is never 0. The critic torso network is 1 dimensional MLP of size 256 with $ELU$ activation on top of it. Both critic encoder and critic torso are applied to the state input and not the action. The output of torso and the action are passed to the head, which firstly applies a $tanh$ activation to the action to scale it in $[-1,1]$ interval, then concatenates both scaled action and torso output. This concatenated output is then passed through a 3-dimensional MLP with sizes $[256, 256, 1]$ with $ELU$ activations applied to all layers except the last one. This produces the Q-function representation, $Q(s,a)$. The critic network is not used for the offline expert cloning and \textsc{DAgger} experiments.

To train experts for \textit{Humanoid Run} and \textit{Humanoid Walls} tasks, we use MPO~\citep{abdolmaleki2018maximum} algorithm with default hyperparamertes. For \textit{Insert Peg} experiments, we use VMPO~\citep{song2019vmpo} algorithm since we found that MPO~\citep{abdolmaleki2018maximum} failed to train. We use the same architecture as described above and use default hyperparameters from VMPO original paper.

\section{Offline policy cloning experiment details}
\label{sec_ap:offline_policy_cloning}

For each task, we train expert policies till convergence. We use MPO algorithm~\cite{abdolmaleki2018maximum} for \textit{Humanoid} tasks and VMPO algorithm~\cite{song2019vmpo} for \textit{Insert Peg} task, as we found MPO was unable to learn on this task.

The policies are represented by the Gaussian distribution $\pi_{E}(\cdot | s) = \mathcal{N}(\mu_{E}(s), \sigma(s))$.
We create datasets as in eqn.~\eqref{eq:bc} using pre-trained experts with a different number of expert trajectories. To asses the sensitivity of different methods to the expert noise, when constructing a dataset, the expert action is drawn according to Gaussian distribution with a fixed variance, i.e.
\begin{equation}
    \label{eq:app_expert_noise}
    a \sim \mathcal{N}(\mu_{E}(s), \sigma_{E}),
\end{equation}
where $\sigma_{E}$ is the fixed amount of expert noise. We consider 4 different levels of $\sigma_{E}$: \textbf{Deterministic}, meaning that we unroll the expert trajectories using only the mean $\mu_{E}$, \textbf{Low}: $\sigma_{E}=0.2$, \textbf{Medium}: $\sigma_{E}=0.5$, \textbf{High}: $\sigma_{E}=1.0$. We also tried values in-between, but did not found a qualitative difference. We also tried values above $\sigma_{E}=1.0$, but the performance for these ones was almost zero. In all the experiments, we use $\sigma_{E} = 0.2$. We provide additional ablation over different levels of expert noise $\sigma_{E}$ in Appendix~\ref{sec_app:apc_teacher_noise}. 

We unroll the expert trajectories by chunks containing $10$ time steps each and put it in a dataset. We use Reverb (from ACME~\citep{hoffman2020acme}) backend for this. A full trajectory for a \textit{Humanoid Run} task corresponds to $1000$ time steps which corresponds to $25$ seconds of control time with a control discretization of $0.025$ seconds. A full trajectory for a \textit{Insert Peg} task corresponds to $1000$ time steps which corresponds to $10$ seconds of control time with a control discretization of $0.1$ seconds. A full trajectory for the \textit{Humanoid Walls} task corresponds to 2000-2500 time steps. This variation is due to potential early stopping of the task execution (in case if the agent falls down). The discretization for the control is $0.03$ seconds and maximum episode length is $45$ seconds.

For each task, we construct datasets containing 1, 2, 3, 5, 10, 20, 50, 100, 200, 500, 1000 trajectories. For \textit{Insert Peg} task, we also create datasets containing 2000, 3000, 5000 and 10000 trajectories, as we found this task requiring more data to be able to be learned.

When evaluating the method, in order to analyze the noise robustness of the student policy is trained via BC, $\pi(\cdot|s) = \mathcal{N}(\mu(s),\sigma(s))$, we evaluate it by executing the action drawn from a Gaussian with a fixed variance, i.e.
\begin{equation}
    \label{eq:app_student_noise}
    a \sim \mathcal{N}(\mu(s), \sigma),
\end{equation}
where $\sigma$ is the fixed amount of student noise. We tried similar value for $\sigma$ as in case of expert noise $\sigma_{E}$. In all the experiments below we use $\sigma=0.2$. We provide additional ablation over these values in Appendix~\ref{sec_app:apc_teacher_noise}.

To train offline expert cloning methods we rely on Algorithm~\ref{alg:apc} as the main algorithm for all the methods, where we remove additional action or/and state augmentations for Naive ABC and BC. For all the experiments we use learning rate $\alpha=0.0001$, number of augmented samples $M=10$, batch size $L=64$. We did try different values for these parameters and found it made no difference on the performance and final experiments outcome, so we fixed one set of parameters for the simplicity of the experimentation.

For every method and task, we tune method-specific hyperparameters. It corresponds to tuning state noise perturbation variance $\sigma_s^2$ from eqn.~\eqref{eq:gaussian_perturbation} for APC and Naive ABC (we do not need to tune it for BC as we do not use any data augmentation in case of BC). The considered range for this parameter is $[0.0001, 0.001, 0.01, 0.05, 0.1, 1.0, 10.0]$. For APC, we found that $\sigma_{s}=0.1$ worked best for \textit{Humanoid Run} and \textit{Insert Peg}, and $\sigma_{s}=1.0$ worked best for \textit{Humanod Walls} task. For Naive ABC, these values are: $\sigma_{s}=0.001$ for \textit{Humanoid Run}, $\sigma_{s}=0.01$ for \textit{Insert Peg} and $\sigma_{s}=0.0001$ for \textit{Humanoid Walls} tasks. We select these parmameters based on the validation set performance in the procedure described in the next paragraph. We use the same values of $\sigma_{s}$ for vision-based augmentation variants as the original method to which these vision-based augmentations are applied. For example, if the method is APC with image, it means that we use $\sigma_{s}$ for APC in this experiment.

We train all the offline expert cloning methods till convergence (maximum 20M iterations) for each experiment / task / method variant. Each iteration corresponds to applying gradients to the batch of 64 trajectories, each containing 10 time steps. We evaluate each model using 150 random environment instantiations. We noticed that when we train models offline, at convergence there are small variations in performance among subsequent models. Therefore, we use early stopping to select for the best model for each experiment. In order to do that, we use a validation set of separate 50 random environment instantiations and we select the best model based on the average performance among these 50 instantiations. We use the same early stopping procedure to select for the best hyperparameter.

\subsection{Expert compression: details}
\label{sec_app:expert_compression_details}

We consider \emph{expert compression} setting as discussed in Section~\ref{sec:expert_compression}, where a student policy has smaller parameters than the expert. This setting often occurs in situations where there are computation constraints (memory, etc.) on the system which would be used on the student as in \citep{parisotto2021efficient}. To study the APC data efficiency in this setting, we consider different sizes of the student network torso, where $[256, 256, 256]$ corresponds to the original network size. In particular, we consider following sizes: $[256], [256, 64], [256, 256]$. We consider additional network sizes and provide additional ablations in Appendix~\ref{sec_app:small_network_add_results}.

\subsection{Learning from privileged experts: details}
\label{sec_app:priveledged_details}

We consider a scenario where the expert has access to the \emph{privileged information} whereas student does not, as discussed in Section~\ref{sec:privileged_expert}. Typically, in such a scenario, it is easier to train the expert than the student, but training a student with a restricted observations is more preferable in an application.

To study the impact of APC in this scenario, we train the expert on \textit{Insert Peg} task where the full state contains common information (proprioception, sword position and orientation) and privileged information of the target position. The student is then trained on the common observations and on vision-based input through the second person camera which replaces privileged information. The latter setup is similar in spirit to \citep{laskin2020reinforcement}. The student network with additional vision observations therefore has an additional visual input encoder as described in Appendix~\ref{sec:agent_architecture}. It encodes non-vision observations with simple concatenation and concatenates the result with the vision embedding. It is then passed through the same torso and head networks as original expert. In this regime, the student does not know about the target position and needs to infer it from vision-based observations.

\section{APC as subroutine: experimental details}
\label{sec_app:apc_as_sub_details}


\subsection{\textsc{DAgger} experiment details}
\label{sec_app:dagger_details}

For each task, we train expert policies to convergence using the MPO algorithm~\cite{abdolmaleki2018maximum}. Each expert is represented by a Gaussian policy, see Appendix~\ref{sec:agent_architecture}. Throughout the experiment, we use the replay buffer of size $1e6$ where each element corresponds to 10-step trajectory, implemented using Reverb (from ACME~\citep{hoffman2020acme}). We use the actor-learning architecture, with 1 actor and 1 learner, where the actor focuses on unrolling current policy and on collecting the data, whereas the learner samples the trajectories from the replay buffer and applies gradient updates on the parameters. When doing so, we control a relative rate of acting / learning via rate limiters as described in \cite{hoffman2020acme} such that for each time step in the trajectory, we apply in average 10 gradient updates. This allows us to be very data efficient and get the full power from the data augmentation technique. In order to achieve it, we set the samples per request (SPI) parameter of the rate limiter to be $T * B * 10$, where $T=10$ is the trajectory length (sample from a replay buffer), $B$ is the batch size ($256$ for Run and $32$ for Walls). When sampling from the replay buffer, we use uniform sampling strategy. When the replay buffer is full, the old data is removed using FIFO-strategy.

For each method and each domain, we run the experiment with 3 random seeds. Normally, in \textsc{DAgger}, the parameter $\beta$ of mixing the experience between the student and an expert, should decrease to $0$ throughout the learning. For simplicity of experimentation, we used fixed values. We report the results using $\beta=0$ and $\beta=0.3$, but we also experimented with values $\beta=0.1, 0.2, 0.4, 0.5$. We found that our chosen values provided most of the qualitative information. The values of state perturbation noise for APC are: $\sigma_{s}=0.1$ for Run and $\sigma_{s}=1.0$ for Walls task. For Naive ABC, these values are: $\sigma_{s}=0.01$ for Run and $\sigma_{s}=0.001$ for Walls tasks. The values which we tried are: $[0.00001, 0.0001, 0.001, 0.01, 0.1, 1.0, 10.0]$.

When collecting the data, we use a mixture of student and expert, which are represented as stochastic policies via Gaussian distributions. For evaluation, we used their deterministic versions, by unrolling only the mean actions.

To train policies via \textsc{DAgger}, we used analytical cross-entropy between expert and student instead of log probability of student on expert mean actions, as we found that it worked better in practice. We provide qualitative comparison in Appendix~\ref{sec:qual_comp}.

\subsection{Kickstarting experiment details}
\label{sec_app:kickstarting_details}

For each task, we train expert policies to convergence using the MPO algorithm~\cite{abdolmaleki2018maximum}. Since in the kickstarting we are interested in outperforming sub-optimal expert, we select experts which achieve around 50 \% of optimal performance on each task. Each expert is represented by a Gaussian policy, see Appendix~\ref{sec:agent_architecture}. We run experiments using a distributed setup with 64 actors and 1 learner, which queries the batches of trajectories (each containing 10 time steps) from a replay buffer of size $1e6$. We use Reverb (from ACME~\citep{hoffman2020acme} as a backend. Batch size is $256$ for Run and $32$ for Walls. We run the sweep over $\lambda$ parameter from eqn. (6) from the main paper. As opposed to \textsc{DAgger}, we do not use the rate-limiter to control the relative ratio between acting and learning as we found that kickstarting in such a regime was unstable. We found that $\lambda=0.0001$ worked best for Run, whereas $\lambda=0.01$ worked best for Walls. The values we tried are: $[0.0001, 0.001, 0.01, 0.1, 1.0, 10.0]$. We found that for higher values of $\lambda$, the learning was faster but the resulting policy did not outperform the expert. On top of running BC methods, we also report the performance of MPO~\cite{abdolmaleki2018maximum} learning from scratch on the task of interest. The values of state perturbation noise for APC are: $\sigma_{s}=0.01$ for Run and $\sigma_{s}=0.01$ for Walls task. For Naive ABC, these values are: $\sigma_{s}=0.00001$ for Run and $\sigma_{s}=0.0001$ for Walls tasks. The values which we tried are: $[0.00001, 0.0001, 0.001, 0.01, 0.1, 1.0, 10.0]$. On top of that, when using the MPO algorithm during kicktasrting, we modify MPO-specific parameters $\epsilon_{\mu}$ and $\epsilon_{\Sigma}$ to $\epsilon_{\mu} = 0.05$ and $\epsilon_{\Sigma} = 0.001$ as we found that using higher values for M-step constraints led to better kickstarting performance. When we apply image-augmentations for kickstarting, we only apply it on the student policy and not on student $Q$-function. Empirically, we found that adding image augmentations for $Q$ function inputs led to worse performance.

\subsection{Plotting details}

When we plot the results in Figure~\ref{fig:dagger_results}, Figure~\ref{fig:kickstarting_results}, Figure~\ref{fig:dagger_objective_00} and Figure~\ref{fig:dagger_objective_03}
we use the following method. For each independent task, method and independent run (seed), we split the data into bins, each containing 10\% of the data. Then, in each bin, the performance is averaged as well as the 95\% confidence interval is calculated. We then report these values in the figure.

\section{Short trajectories experiment}
\label{sec_app:short_traj_experiment}

\begin{figure}[t]
    \centering
    \includegraphics[width=\textwidth]{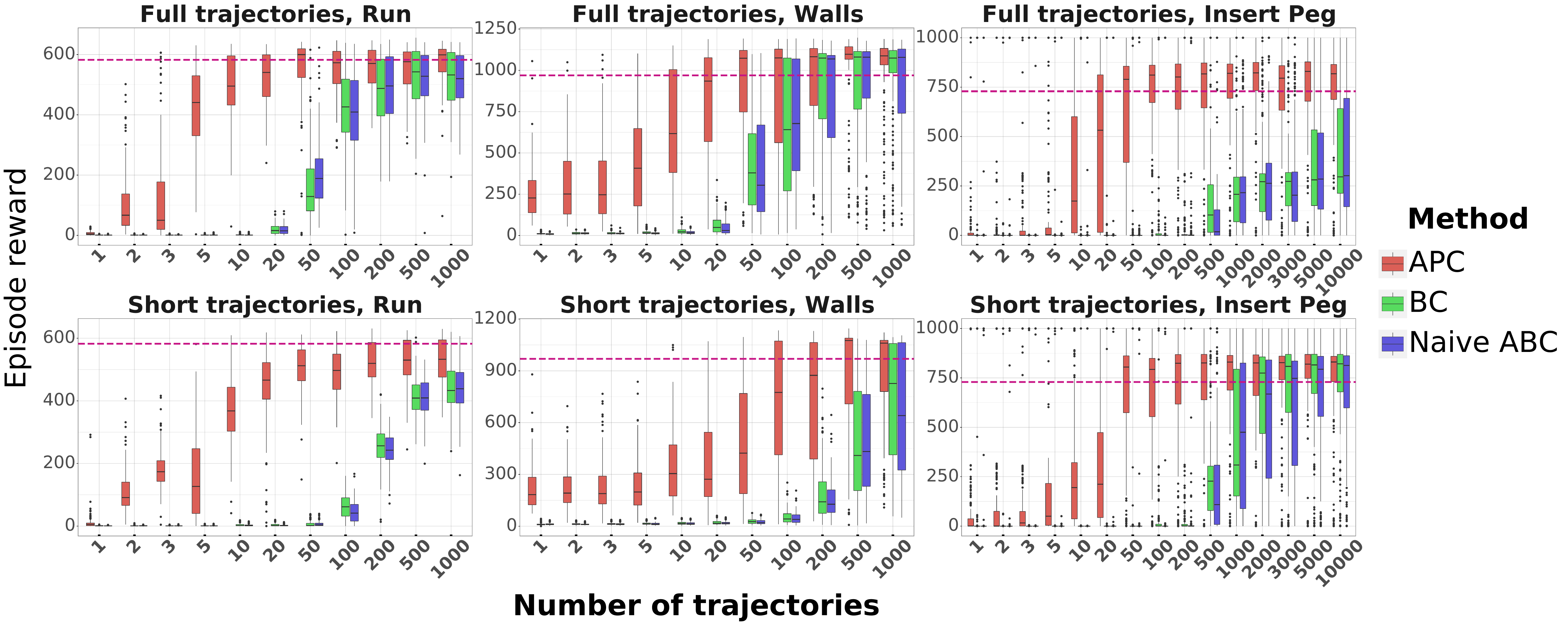}
    \caption{\textbf{Offline expert cloning} results with different number of trajectories for APC, BC and Naive ABC on \textit{Humanoid Run}, \textit{Humanoid Walls} and \textit{Insert Peg} tasks represented by columns. First row corresponds to the full trajectory case, as reported in Figure~\ref{fig:bc_results}. Second row corresponds to the case of short trajectories, where the dataset contain one full and a given number of short trajectories. The X-axis represents the number of trajectories, the Y-axis corresponds to the episodic reward across 150 independent evaluations.}  
    \label{fig:full_and_short_bc_traj}
\end{figure}

In this section we present additional results to the ones presented in Section~\ref{sec:results}. We discussed that we construct the dataset of expert trajectories containing full trjaectories (1000 timesteps for \textit{Humanoid Run} and \textit{Insert Peg} and around 2000 timesteps for \textit{Humanoid Walls}). It corresponded to a simple unroll of the expert policy on the original environments. In addition to that, we create datasets which contain only one full trajectories and a given number of short trajectories, where each short trajectory contains only 200 timesteps starting from the initial state. The reason for this experiment is to study the ability of different offline expert cloning methods for a more data restricted setup. Note that such a scenario can occur in practice when dealing with realistic robots, where dataset can contain a lot of successful trajectories, but the trajectories can be short, because a robot can fail at some points of time.

We present results in Figure~\ref{fig:full_and_short_bc_traj}, where we duplicate the results for full trajectories and add additional results for short trajectories. We observe that APC performs well in all the cases, whereas BC and Naive ABC performance degrade on \textit{Humanoid Run} and on \textit{Humanoid Walls}. What is interesting, however, is that these methods perform better on \textit{Insert Peg} scenario with short trajectories (but still worse than APC). The reason for is due to the fact that in \textit{Insert Peg}, the rewarding state corresponds to a situation where arm inserted a sword into a hole and does not move (and episode is not finished until 1000 time steps had elapsed). Therefore, long trajectories of expert policies will contain a lot of such states and actions, therefore having a limited diversity. In case of short trajectories, the relative ratio of states preceding this final state is much higher. Interestingly, APC still performs well in both scenarios.

\section{Ablations}

\begin{figure}
     \centering
    \includegraphics[width=\textwidth]{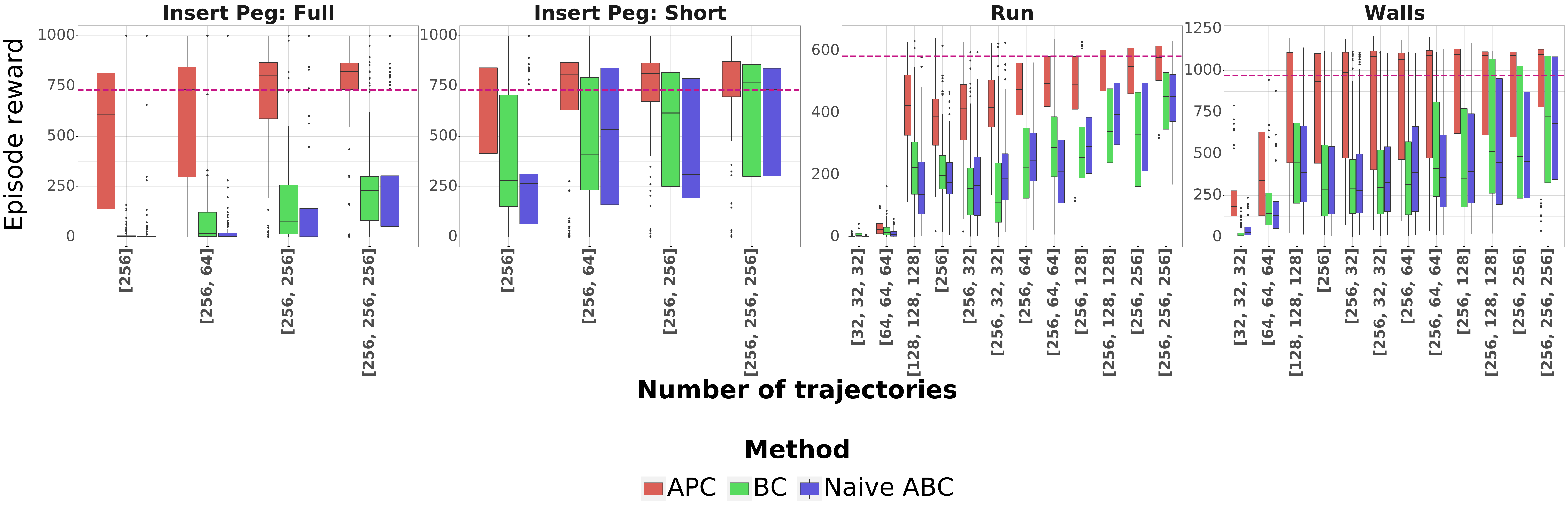}
    \caption{\emph{Teacher compression} results with additional student torso architecture sizes.}
    \label{fig:bc_small_networks_sizes}
\end{figure}

\subsection{\textit{Expert compression} additional results}
\label{sec_app:small_network_add_results}

In this section we present additional results for \textit{expert compression} experiment from Section~\ref{sec:expert_compression}.

Firstly, we present an ablation over different network sizes in Figure~\ref{fig:bc_small_networks_sizes}. We see that generally APC degrades much less than other methods when we decrease the student network size.

\begin{figure}[ht]
     \centering
    \includegraphics[width=\textwidth]{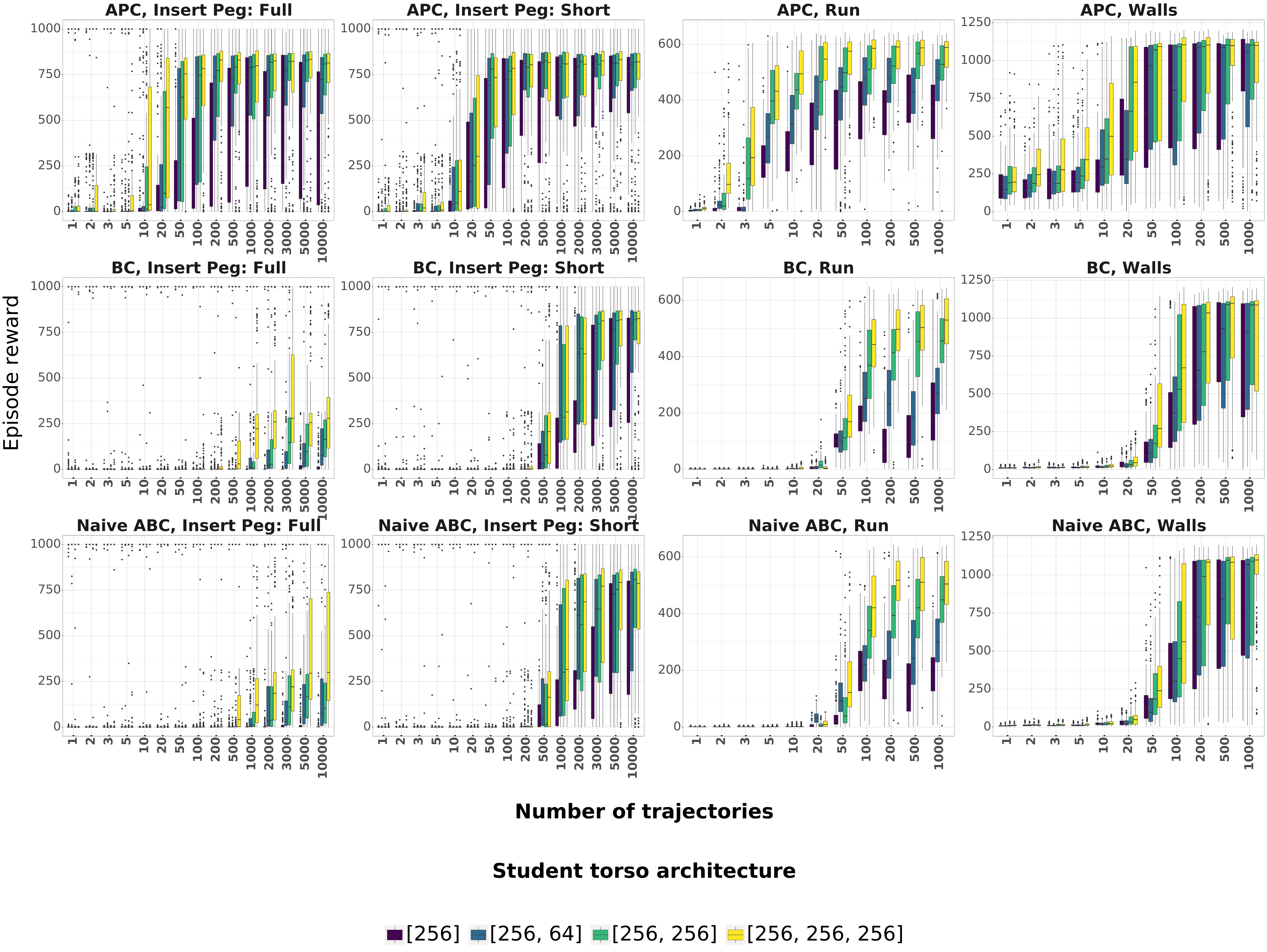}
    \caption{\emph{Teacher compression} with all the methods and tasks. For \textit{Insert Peg}, we consider two setups, with full and short trajectories in the dataset. See Appendix~\ref{sec_app:short_traj_experiment}}
    \label{fig:bc_small_networks_complete}
\end{figure}

Secondly, as an additional to the results in Figure~\ref{fig:bc_small_networks_num_traj}, we present results for all the tasks and all the methods on Figure~\ref{fig:bc_small_networks_complete}

\subsection{Learning from privileged experts: additional results}

\begin{figure}[ht]
     \centering
    \includegraphics[width=\textwidth]{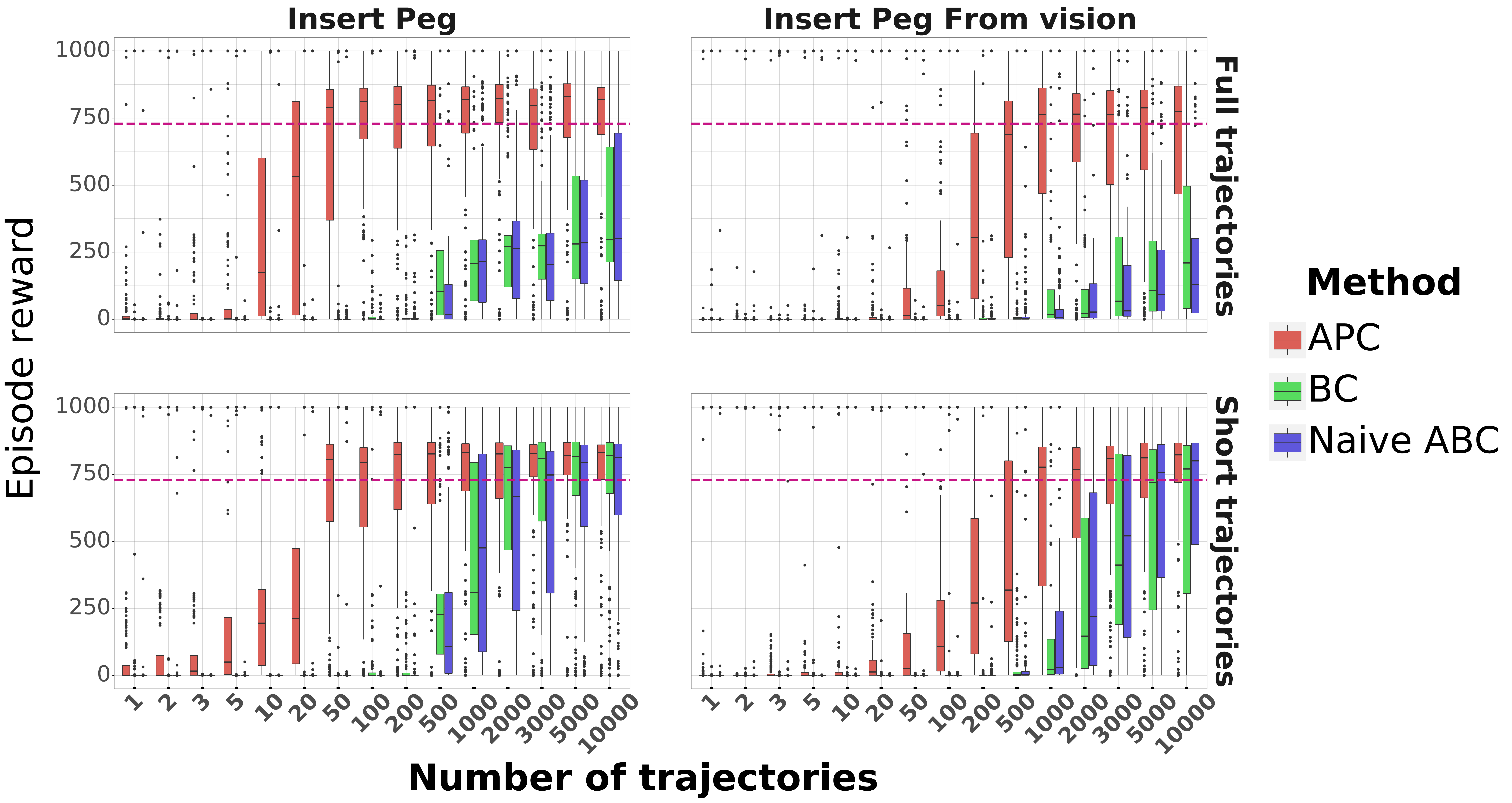}
    \caption{Learning from privileged experts, additional results. On top of the considered results in the main paper, we add additional results with short trajectories.}
    \label{fig:priv_experts_add_results}
\end{figure}

We present additional results to the experiment presented in Section~\ref{sec:privileged_expert} where we also consider \textit{Insert Peg} tasks where dataset contains only short trajectories as we have seen in Appendix~\ref{sec_app:short_traj_experiment} that all methods performed better on this task with short trajectories. The results are given in Figure~\ref{fig:priv_experts_add_results}.

\subsection{APC expert noise sensitivity}
\label{sec_app:apc_teacher_noise}

\begin{figure}[ht]
    \includegraphics[width=\textwidth]{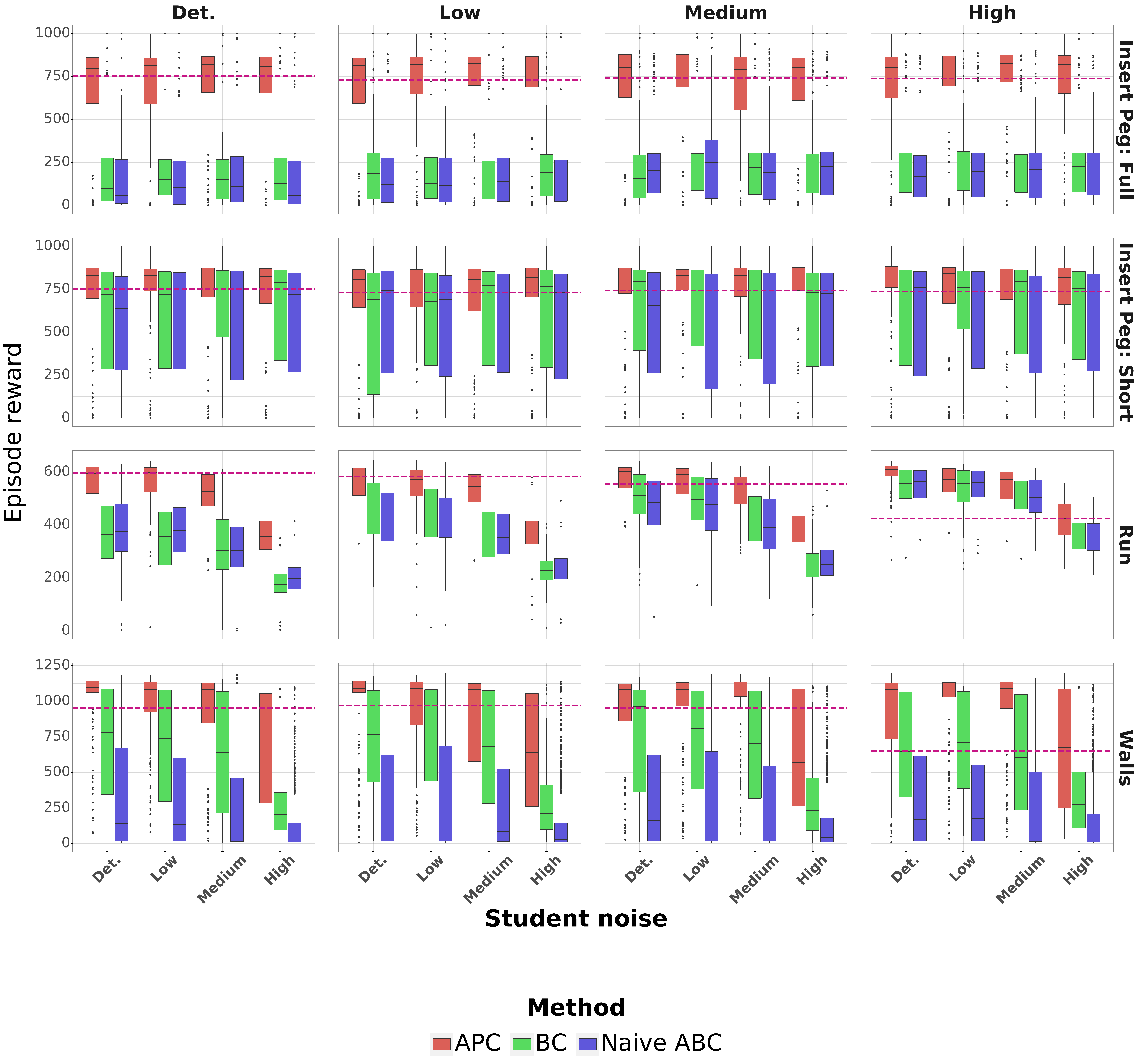}
    \caption{\textbf{Noise sensitivity results.} We consider 4 levels of noise for student and expert: \textbf{Deterministic}, which uses the Gaussian mean for the action, \textbf{Low}, is the noise $\sigma=0.2$, \textbf{Medium} $\sigma=0.5$ and \textbf{High} $\sigma=1.0$. Each column corresponds to a different level of expert noise. Each row represents different task. For \textit{Insert peg} we consider scenario with full and short trajectories, see Appendix~\ref{sec_app:short_traj_experiment}. For \textit{Humanoid Run} and \textit{Humanoid Walls} tasks, we use 100 trajectories dataset, whereas for \textit{Insert Peg}, we use dataset with 2000 trajectories, as this task is much less data efficient than others. X-axis corresponds to a different level of student noise. Y-axis corresponds to the episodic reward with 150 independent evaluations. The legend denotes a method and a row corresponds to a task. The pink dashed line indicate average expert performance.} 
    \label{fig:bc_expert_noise_sensitivity}
\end{figure}

In this section we present additional ablations on the sensitivity of APC, BC and Naive ABC to different values of expert noise $\sigma_{E}$ from eqn.~\eqref{eq:app_expert_noise} and student noise $\sigma$ from eqn.~\eqref{eq:app_expert_noise}. We consider four different levels of noise as discussed in Appendix~\ref{sec_ap:offline_policy_cloning}. The results are given in Figure~\ref{fig:bc_expert_noise_sensitivity}.  For \textit{Humanoid Run} and \textit{Humanoid Walls} tasks, we use 100 trajectories dataset, whereas for \textit{Insert Peg}, we use dataset with 2000 trajectories, as this task is much less data efficient than others. Moreover, for \textit{Insert peg} we consider scenario with full and short trajectories, whereas for \textit{Humanoid} tasks we consider only full trajectories. We see that overall, APC provides more robust policies for different amounts of expert and student noise. For expert noise sensitivity, note that over different columns, the APC performance degrades much less than for BC and Naive ABC. Moreover, for each column, observing the change of the student noise level (from low to high), we see that performance degrades for all the methods, but much less for APC. Therefore, APC seems to provide more action-noise robust policies. We see that BC and Naive ABC perform similarly in terms of robustness. Finally, what is interesting, APC generally observes much less variance in performance when varying the noise levels compared to BC and Naive ABC

\subsection{APC and ABC state noise ablations}

\begin{figure}[ht]
    \centering
    \includegraphics[width=\textwidth]{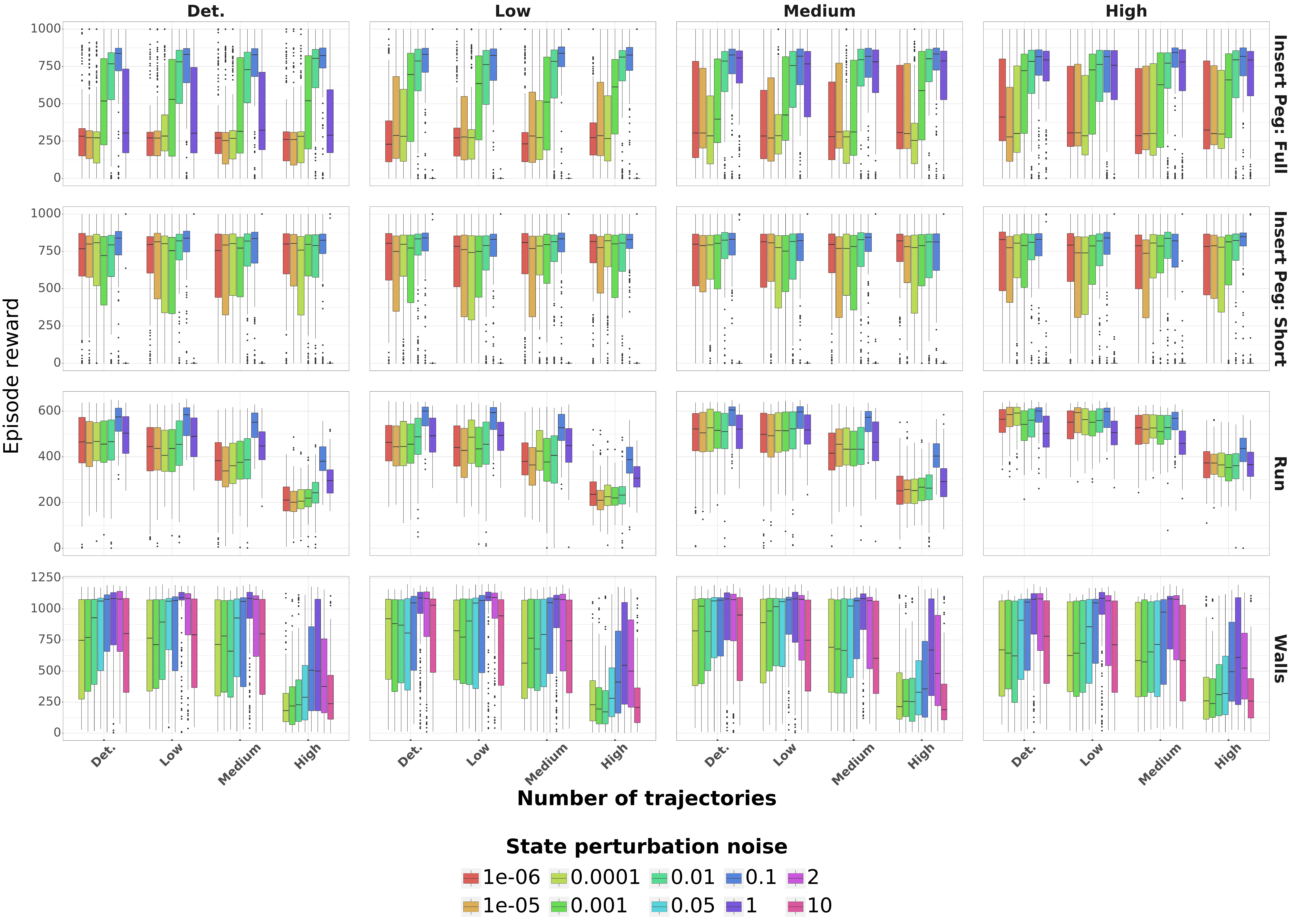}
    \caption{\textbf{State perturbation noise sensitivity for APC.} In this plot we represent the APC method trained on 100 full trajectories sampled under different level of expert noise which is represented by different columns. On the X-axis is the different level of a student noise at evaluation time. The legend denotes different levels of a state perturbation noise $\sigma_{s}$ from the eqn. (7) from the main paper. Y-axis corresponds to the episodic reward with 150 independent evaluations.} 
    \label{fig:bc-state_noise_ablation}
\end{figure}

\begin{figure}[ht]
    \centering
    \includegraphics[width=\textwidth]{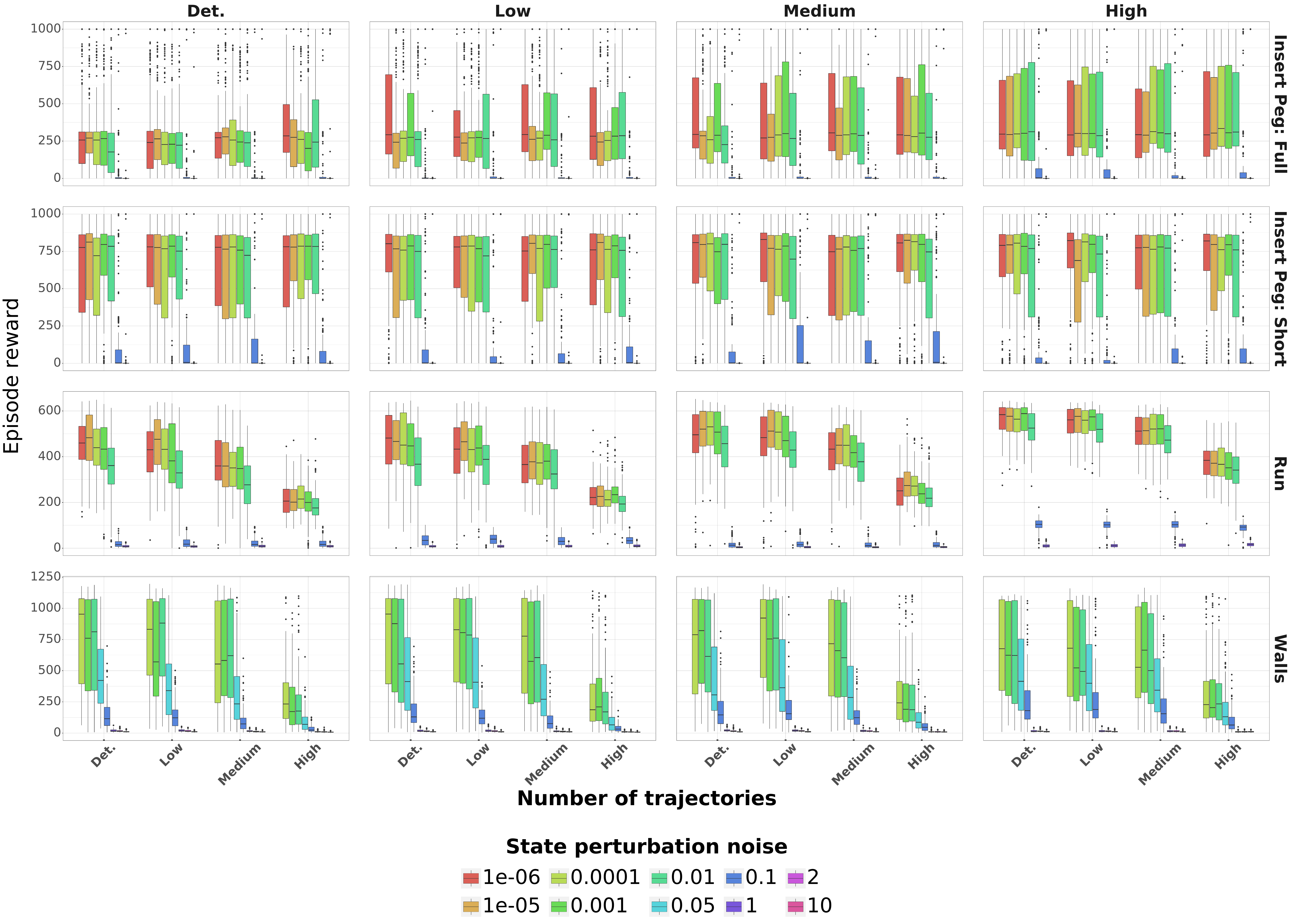}
    \caption{\textbf{State perturbation noise sensitivity for Naive ABC}. In this plot we represent the Naive ABC method trained on 100 full trajectories sampled under different level of expert noise which is represented by different columns. On the X-axis is the different level of a student noise at evaluation time. The legend denotes different levels of a state perturbation noise $\sigma_{s}$ from the eqn. (7) from the main paper. Y-axis corresponds to the episodic reward with 150 independent evaluations.} 
    \label{fig:abc-state_noise_ablation}
\end{figure}

In this section we provide additional ablations for the state-noise perturbation level $\sigma_{s}$ from the eqn. (7) from the main paper. In Figure~\ref{fig:bc-state_noise_ablation}, we show the results for APC, whereas in Figure~\ref{fig:abc-state_noise_ablation}, we show the results for Naive ABC. We see that there is a sweet spot for the state perturbation noise level.

\subsection{Additional comparisons for Walls task}
\label{sec_app:walls_additional_results}

\begin{figure}[ht]
    \centering
    \includegraphics[width=\textwidth]{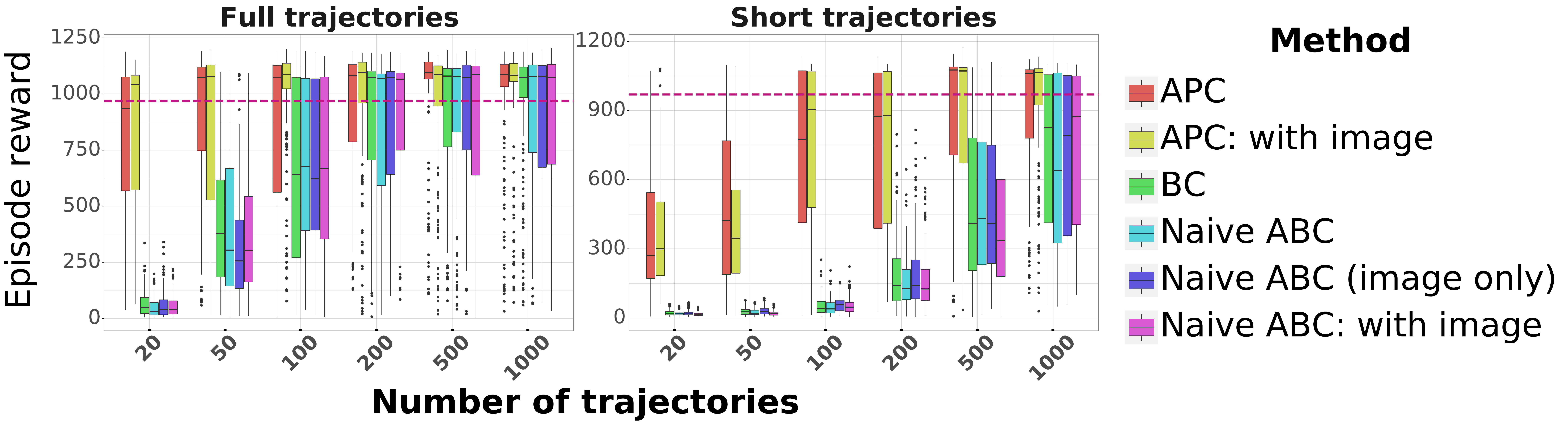}
    \caption{\textbf{Additional behavioral cloning results on Walls tasks with additional methods added.} X-axis corresponds to a number of trajectories used in each of the dataset. The Y-axis corresponds to the episodic reward with 150 random evaluations. The pink dashed line indicate average (among the same 150 independent evaluations) expert performance. The legend describes a method which is used. The plot o the left depicts the performance of offline policy cloning with using full trajectories from the expert, whereas the plot on the right represents the experiment with short trajectories. See Appendix~\ref{sec_app:short_traj_experiment} for more details.
    } 
    \label{fig:bc_vision_results}
\end{figure}

In Figure~\ref{fig:bc_vision_results}, we provide additional results for behavioral cloning experiment on Walls task where we try different variants of APC and Naive ABC with additional image-based augmentation as described in the main paper.

\subsection{Objective functions comparison for \textsc{DAgger}}
\label{sec:qual_comp}

In Figure~\ref{fig:dagger_objective_00} and in Figure~\ref{fig:dagger_objective_03}, we provide ablations over different objectives for \textsc{DAgger} with $\beta=0.0$ and $\beta=0.3$ correspondingly. We see that overall, training with cross-entropy leads to better results than with log prob on the mean action, especially when $\beta=0.0$.

\begin{figure}[ht]
    \includegraphics[width=\textwidth]{new_images/bc-dagger-objective_00.pdf}
    \caption{\textbf{\textsc{DAgger} objective sweep with $\beta=0.0$.} On the X-axis we report the number of environment steps. On the Y-axis we report averaged across 3 seeds episodic reward achieved by the student. Shaded area corresponds to confidence intervals. For a Run task, the confidence intervals are small, so they are not visible. In solid line we report the performance when training using the cross-entropy. In dashed line, we report the performance when training using log probability on the mean action from the expert. All the methods use mean action during evaluation. The black dashed line indicate average (among the same 150 independent evaluations) expert performance for the given expert noise level.} 
    \label{fig:dagger_objective_00}
\end{figure}

\begin{figure}[ht]
    \includegraphics[width=\textwidth]{new_images/bc-dagger-objective_03.pdf}
    \caption{\textbf{\textsc{DAgger} objective sweep with $\beta=0.3$.} On the X-axis we report the number of environment steps. On the Y-axis we report averaged across 3 seeds episodic reward achieved by the student. Shaded area corresponds to confidence intervals. For a Run task, the confidence intervals are small, so they are not visible. In solid line we report the performance when training using the cross-entropy. In dashed line, we report the performance when training using log probability on the mean action from the expert. All the methods use mean action during evaluation. The black dashed line indicate average (among the same 150 independent evaluations) expert performance for the given expert noise level.} 
    \label{fig:dagger_objective_03}
\end{figure}

\end{document}


\maketitle

\section{Experimental details}

\begin{figure*}
     \centering
    \includegraphics[width=\textwidth]{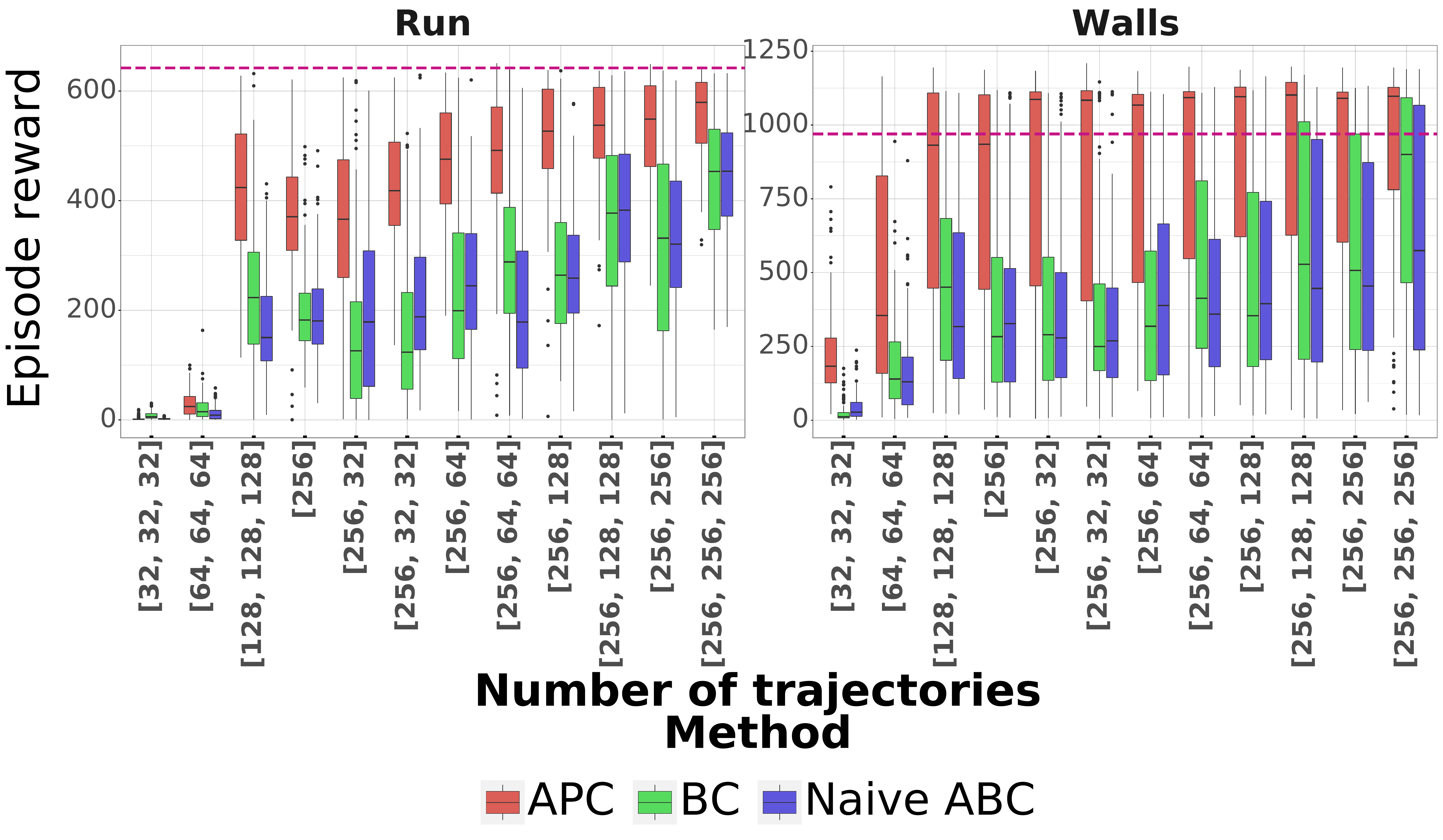}
    \caption{\emph{Teacher compression} results with all the considered architecture sizes.}
    \label{fig:bc_small_networks_num_traj}
\end{figure*}

\subsection{TO ADD for vision tasks}

For the APC method, we rely on Algorithm~\ref{alg:apc}. For baselines, we consider BC algorithm from eqn.~\eqref{eq:bc} as well as a simple modification of BC, where we apply, similar to APC, state perturbations as in eqn.~\eqref{eq:gaussian_perturbation} and eqn.~\eqref{eq:state_perturbation}, but we do not produce a new action from the expert. We call this approach Naive Augmented Behavior Cloning (Naive ABC) which essentially corresponds to robustification of the student policies with respect to state perturbation and is similar in spirit to standard data-augmentation approaches. 
For vision-based tasks, we consider random crop augmentations of size 48x48 (downsampled from the input image of 64x64), similar to \cite{laskin2020reinforcement}. When the image augmentations are used we add "with image" to the method name. On top of that, we consider a variant, where only image augmentation is used, which we call Naive ABC (image only). For all methods, as an action in the objective from eqn.~\eqref{eq:bc}, we use an expert mean $\mu_{E}(s)$.

\subsection{TO ADD MAKE SURE IT IS INCORPORATED}
We train all approaches to convergence (300K learning iterations on Walls and 13M learning iterations on Run). Each learning iteration corresponds applying gradients to 64 trajectories, each containing 10 time steps. After each learning iteration, we evaluate the policy on both a validation set (50 random instances of the environment) and a test set (150 random instances of the environments). We apply early-stopping based on the validation set performance to select the best model and report corresponding performance on the test set. For more details, please refer to Section 1.3 in Supplementary Material.

\subsection{Environment details}
In this work we consider two environments from DeepMind Control suite~\citep{tunyasuvunakool2020dm_control}: \textbf{Run} task with \textbf{Humanoid} body and Walls task (original name is "run through a corridor") with \textbf{Humanoid} body (original body was \textbf{CMU Humanoid}). In \textbf{Run} task, an agent must run at a target speed and gets reward which is proportional to the inverse distance between its current speed and the target speed. The agent receives proprioceptive observations only, which fully describe the state. In \textbf{Walls} task, an agent must run through the corridor and avoid the emerging walls. It receives reward which is proportional to the forward speed (through the corridor), thus incentivising it to run as fast as possible. It receives proprioceptive observations as well as the image of size 64x64 from the ego-centric camera. Action dimension is equal to 21. For more details, check \cite{tunyasuvunakool2020dm_control}.

\subsection{Agent architecture}
\label{sec:agent_architecture}
For all the experiments, we use the same agent architecture. The agent has two separate networks: actor (policy) and critic (Q-function). Both networks are split into 3 components: encoder, torso and head. Encoders for actor and critic are separate but have the same architecture. For state-only (no vision) observations, encoder corresponds to a simple concatenations of all the observations. For the vision input, it divides each pixel by $255$ and then applies a 3-layer ResNet of sizes $(16, 32, 32)$ with $ELU$ activations followed by a linear layer of size 256 and $ELU$ activation. The resulting output is then concatenated together with state-based input. For actor network, torso corresponds to a 3 dimensional MLP, each hidden layer of size $256$ with activation $ELU$ applied at the end of each hidden layer. The output of actor torso is then passed to the actor head network, which applies a linear layer (without activation) with output size equal to $N_{a} * 2$, where $N_{a}$ is the action dimension (21 in our case). It produces the actor mean $\mu$ and log-variance: $\log \tilde{\sigma}$. Then, the variance of the actor is calculated as
\begin{equation*}
    \sigma = soft plus (\log \tilde{\sigma})  + \sigma_{min},
\end{equation*}
where $\sigma_{min} = 0.0001$. That would encode the Gaussian policy $\pi(\cdot | s) = \mathcal{N}(\mu(s) | \sigma(s))$. This parameterisation insures that the variance is never 0. The critic torso network is 1 dimensional MLP of size 256 with $ELU$ activation on top of it. Both critic encoder and critic torso are applied to the state input and not the action. The output of torso and the action are passed to the head, which firstly applies a $tanh$ activation to the action to scale it in $[-1,1]$ interval, then concatenates both scaled action and torso output. This concatenated output is then passed through a 3-dimensional MLP with sizes $[256, 256, 1]$ with $ELU$ activations applied to all layers except the last one. This produces the Q-function representation, $Q(s,a)$. The critic network is not used for the Behavioral Cloning and \textsc{DAgger} experiments.

To train experts and agents in kickstarting experiment, we use MPO~\citep{abdolmaleki2018maximum} algorithm. As for MPO-specific parameters, we used $\epsilon_{\mu} = 0.05$ and $\epsilon_{\Sigma} = 0.001$ as we found that using higher values for M-step constraints led to better kickstarting performance. For both tasks, we train MPO agents till convergence. In order to test the ability of the kickstarting method to outperform the sub-optimal experts, we keep partially trained expert policies by saving corresponding checkpoints. Then, we keep 3 types of experts: \textbf{Low}, the one achieving approximately 25\% of the converged expert performance, \textbf{Medium}, the one achieving around 50 \% of expert performance and \textbf{High}, the one achieving the expert performance.

\subsection{Behavioral cloning experiment details}

For behavioral cloning experiments, for both tasks, we consider converged experts (expert type is \textbf{High}). Given this expert, we produce fixed size datasets (for different number of trajectories), by unrolling the expert with a fixed noise level. Since the expert is a Gaussian policy, $\mathcal{N}(\mu_{E}(s), \sigma_{E}(s))$, we unroll the trajectories with actions sampled with a fixed noise level $\sigma$:
\begin{equation*}
    a_{E} \sim \mathcal{N}(\mu_{E}(s), \sigma)
\end{equation*}
We consider 4 different levels of $\sigma$: \textbf{Deterministic}, meaning that we unroll the expert trajectories using only the mean $\mu_{E}$, \textbf{Low}: $\sigma=0.2$, \textbf{Medium}: $\sigma=0.5$, \textbf{High}: $\sigma=1.0$. We also tried values in-between, but did not found a qualitative difference. We also tried values above $\sigma=1.0$, but the performance for these ones was almost zero. We unroll the expert trajectories by chunks containing $10$ time steps each and put it in a dataset. We use Reverb (from ACME~\citep{hoffman2020acme}) backend for this. A full trajectory for a Run task corresponds to $1000$ time steps which corresponds to $25$ seconds of control time with a control discretization of $0.025$ seconds. A full trajectory for the Walls task corresponds to 2000-2500 time steps. This variation is due to potential early stopping of the task execution (in case if the agent falls down). The discretization for the control is $0.03$ seconds and maximum episode length is $45$ seconds.

For each task, we construct datasets containing 1, 2, 3, 5, 10, 20, 50, 100, 200, 500, 1000 trajectories.

For each of the task and the dataset variant (number of trajectories or/and expert noise level), we train a policy via Behavioral cloning by optimizing the objective from eqn. (3) from the main paper, where the expert action is replaced by the expert mean $\mu_{E}(s)$. More formally, we apply the Algorithm (1) from the main paper where for BC and naive-ABC we either do not apply state perturbation or do not produce a new expert action. For all the experiments we use a learning rate $\alpha=0.0001$, number of augmented samples $M=10$, batch size of $L=64$. We run the experiment for $K=13M$ steps for Run task and for $K=300K$ steps for Walls task. For each learning iteration, we evaluate the model on $50$ random instances of environment, which corresponds to a validation set, and on $150$ random instances of environment corresponding to a test set. We use validation set to select parameters, such as state perturbation noise $\sigma_{s}$ as well as to select among the converged models (as we observed that when the models converged, there were always small variations in performance). We then reported the performance on the test set for all the plots. The values of state perturbation noise for APC are: $\sigma_{s}=0.1$ for Run and $\sigma_{s}=1.0$ for Walls task. For Naive ABC, these values are: $\sigma_{s}=0.001$ for Run and $\sigma_{s}=0.01$ for Walls tasks. The values which we tried are: $[0.0001, 0.001, 0.01, 0.05, 0.1, 1.0, 2.0, 10.0]$. For the experiments using short trajectories, the datasets were built by adding 1 full trajectory (1000 time steps for Run task and around 2000-2500 time steps for Walls tasks) with the rest containing only short trajectories (fixed to 200 time steps).

\subsection{\textsc{DAgger} experiment details}

For \textsc{DAgger}, similarly to BC experiments, we use \textbf{High} type of the expert (converged one). Throughout the experiment, we use the replay buffer of size $1e6$ where each element corresponds to 10-step trajectory, implemented using Reverb (from ACME~\citep{hoffman2020acme}). We use the actor-learning architecture, with 1 actor and 1 learner, where the actor focuses on unrolling current policy and on collecting the data, whereas the learner samples the trajectories from the replay buffer and applies gradient updates on the parameters. When doing so, we control a relative rate of acting / learning via rate limiters as described in \cite{hoffman2020acme} such that for each time step in the trajectory, we apply in average 10 gradient updates. This allows us to be very data efficient and get the full power from the data augmentation technique. In order to achieve it, we set the samples per request (SPI) parameter of the rate limiter to be $T * B * 10$, where $T=10$ is the trajectory length (sample from a replay buffer), $B$ is the batch size ($256$ for Run and $32$ for Walls). When sampling from the replay buffer, we use uniform sampling strategy. When the replay buffer is full, the old data is removed using FIFO-strategy.

For each method and each domain, we run the experiment with 3 random seeds. Normally, in \textsc{DAgger}, the parameter $\beta$ of mixing the experience between the student and an expert, should decrease to $0$ throughout the learning. For simplicity of experimentation, we used fixed values. We report the results using $\beta=0$ and $\beta=0.3$, but we also experimented with values $\beta=0.1, 0.2, 0.4, 0.5$. We found that our chosen values provided most of the qualitative information. The values of state perturbation noise for APC are: $\sigma_{s}=0.1$ for Run and $\sigma_{s}=1.0$ for Walls task. For Naive ABC, these values are: $\sigma_{s}=0.01$ for Run and $\sigma_{s}=0.001$ for Walls tasks. The values which we tried are: $[0.00001, 0.0001, 0.001, 0.01, 0.1, 1.0, 10.0]$.

When collecting the data, we use a mixture of student and expert, which are represented as stochastic policies via Gaussian distributions. For evaluation, we used their deterministic versions, by unrolling only the mean actions.

To train policies via \textsc{DAgger}, we used analytical cross-entropy between expert and student instead of log probability of student on expert mean actions, as we found that it worked better in practice. We provide qualitative comparison in Section~\ref{sec:qual_comp} in Supplementary material.

\subsection{Kickstarting experiment details}

For kickstarting experiment, we use 3 types of expert policy, as described in Section~\ref{sec:agent_architecture} in Supplementary material. We run experiments using a distributed setup with 64 actors and 1 learner, which queries the batches of trajectories (each containing 10 time steps) from a replay buffer of size $1e6$. We use Reverb (from ACME~\citep{hoffman2020acme} as a backend. Batch size is $256$ for Run and $32$ for Walls. We run the sweep over $\lambda$ parameter from eqn. (6) from the main paper. As opposed to \textsc{DAgger}, we do not use the rate-limiter to control the relative ratio between acting and learning as we found that kickstarting in such a regime was unstable. We found that $\lambda=0.0001$ worked best for Run, whereas $\lambda=0.01$ worked best for Walls. The values we tried are: $[0.0001, 0.001, 0.01, 0.1, 1.0, 10.0]$. We found that for higher values of $\lambda$, the learning was faster but the resulting policy did not outperform the expert. On top of running BC methods, we also report the performance of MPO~\cite{abdolmaleki2018maximum} learning from scratch on the task of interest. The values of state perturbation noise for APC are: $\sigma_{s}=0.01$ for Run and $\sigma_{s}=0.01$ for Walls task. For Naive ABC, these values are: $\sigma_{s}=0.00001$ for Run and $\sigma_{s}=0.0001$ for Walls tasks. The values which we tried are: $[0.00001, 0.0001, 0.001, 0.01, 0.1, 1.0, 10.0]$.

\subsection{Computational resources}

For pretraining two expert policies on two tasks, we used in \textbf{total $1$ GPU v100 and 64 4-core CPUs and around 500Gb memory for each run.}

For Behavioral cloning experiments, we used $1$ GPU p100 and 1 CPU with 4 cores and around 64GB of memory for each run. In total we had a sweep over state-noise perturbation for both APC and Naive ABC, where sweep contained 8 experiments, resulting in 16 experiments in total (it would correspond to Figure 1 and Figure 2 from the supplementary material), which should be multiplied by 2 as we use 2 tasks, resulting in \textbf{32 experiments each with 1 p100 GPU, 1 CPU with 4 cores and around 64GB of memory}.

To produce Figure 1 in the main paper, we run 11 (different number of trajectories) * 3 (number of methods) * 2 (full or short trajectories) = 66 experiments. To produce Figure 2 in the main paper, we run 4 (different expert noises) * 3 (number of methods) = 12 experiments. To produce Figure 3 from supplementary material, we ran additionally 6 (different number of trajectories) * 3 (additional methods) * 2 (short or full trajectories) = 36 experiments. On top of that, we need to multiply it by 2 (as we have 2 tasks), resulting in 114 * 2 = 228 experiments. \textbf{Therefore, for behaviour cloning part, we ran in total 260 experiments, each using 1 p100 GPU and 1 CPU with 4 cores and around 64Gb of memory}

For each \textsc{DAgger} experiment, we use 1 v100 GPU, 4 CPU with 4 cores each (for Replay, actor, evaluator, and remover), around 100Gb of memory in total. We conduct 3 (different methods) * 3 (different seeds) * 2 (different beta) * 2 (different tasks) = 36 experiments for non-vision based methods and 3 (different methods) * 3 (different seeds) * 2 (different beta) = 18 experiments for vision based methods. On top of that, we conducted a sweep over $\sigma_{s}$ for APC and Naive ABC and for each task, which leads 84 other experiments. \textbf{In total, we conducted 138 experiments each using v100 GPU, 4 CPU with 4 cores and 100Gb memory.}

For each kickstarting, we used 1 v100 GPU for walls and 1 p100 GPU for run tasks, 74 CPU with 4 cores each (64 for actors, 8 for evaluators, 1 for replay, 1 for remover), around 800Gb of memory in total. We conducted 84 experiments to sweep over $\sigma_{s}$ for APC and Naive ABC, and 36 experiments to sweep over $\lambda$ parameter. On top of that, we conducted in total 3 (different methods) * 3 (different seeds) * 3 (different experts) * 2 (different tasks) = 54 experiments for non-vision based methods and 3 (different methods) * 3 (different seeds) * 3 (different experts) = 27 experiments for vision-based methods. \textbf{In total, it would correspond to 200 experiments, each half of which used v100 GPUs and half of which used p100 GPU, 74 CPU with 4 cores and around 800Gb of memory.}

\subsection{Plotting details}

When we plot the results in Figure 3, Figure 4 from the main paper and in Figure~\ref{fig:dagger_objective_0} and Figure~\ref{fig:dagger_objective_03} in Supplementary material, we use the following method. For each independent task, method and independent run (seed), we split the data into bins, each containing 10\% of the data. Then, in each bin, the performance is averaged as well as the 95\% confidence interval is calculated. We then report these values in the figure.

\section{APC teacher noise sensitivity}

\begin{figure}[h!]
    \includegraphics[width=0.9\linewidth]{images/bc-bc_teacher_noise_sensitivity_new.pdf}
    \caption{\textbf{Noise sensitivity results.} We consider 4 levels of noise for student and expert: \textbf{Deterministic}, which uses the Gaussian mean for the action, \textbf{Low}, is the noise $\sigma=0.2$, \textbf{Medium} $\sigma=0.5$ and \textbf{High} $\sigma=1.0$. Each column corresponds to a different level of expert noise. X-axis corresponds to a different level of student noise. Y-axis corresponds to the episodic reward averaged among 150 independent evaluations. The highest point of the bar corresponds to the mean, whereas the dashed lines indicate the standard deviation. The legend denotes a method and a row corresponds to a task. The pink dashed line indicate average expert performance.} 
    \label{fig:bc_expert_noise_sensitivity}
\end{figure}

\section{APC and ABC state noise ablations}

In this section we provide additional ablations for the state-noise perturbation level $\sigma_{s}$ from the eqn. (7) from the main paper. In Figure~\ref{fig:bc-state_noise_ablation}, we show the results for APC, whereas in Figure~\ref{fig:abc-state_noise_ablation}, we show the results for Naive ABC. We see that there is a sweet spot for the state perturbation noise level.

\begin{figure}[h!]
    \includegraphics[width=0.9\linewidth]{images/bc-state_noise_ablation_new.pdf}
    \caption{\textbf{State perturbation noise sensitivity for APC.} In this plot we represent the APC method trained on 100 full trajectories sampled under different level of expert noise which is represented by different columns. On the X-axis is the different level of a student noise at evaluation time. The legend denotes different levels of a state perturbation noise $\sigma_{s}$ from the eqn. (7) from the main paper. The Y-axis represents the averaged among 150 independent evaluations episode reward.} 
    \label{fig:bc-state_noise_ablation}
\end{figure}

\begin{figure}[h!]
    \includegraphics[width=0.9\linewidth]{images/abc-state_noise_ablation_new.pdf}
    \caption{\textbf{State perturbation noise sensitivity for ABC}. In this plot we represent the APC method trained on 100 full trajectories sampled under different level of expert noise which is represented by different columns. On the X-axis is the different level of a student noise at evaluation time. The legend denotes different levels of a state perturbation noise $\sigma_{s}$ from the eqn. (7) from the main paper. The Y-axis represents the averaged among 150 independent evaluations episode reward.} 
    \label{fig:abc-state_noise_ablation}
\end{figure}

\section{Additional comparisons for Walls task}

In Figure~\ref{fig:bc_vision_results}, we provide additional results for behavioral cloning experiment on Walls task where we try different variants of APC and Naive ABC with additional image-based augmentation as described in the main paper.

\begin{figure}[h!]
    \includegraphics[width=0.9\linewidth]{images/bc-num_traj_only_walls.pdf}
    \caption{\textbf{Additional behavioral cloning results on Walls tasks with additional methods added.} X-axis corresponds to a number of trajectories used in each of the dataset. The Y-axis corresponds to the episodic reward. The high point of the bar plot corresponds to the average value (averaged across 150 independent evaluations) and the dashed lines indicate the standard deviation. The pink dashed line indicate average (among the same 150 independent evaluations) expert performance. The legend describes a method which is used. On the plot on the left depicts a standard BC experiment, where dataset contains a specified number of full trajectories from the expert. The plot on the right illustrates the experiment, where a dataset contains 1 full trajectories and the rest are the short ones, containing only 200 timesteps each.
    } 
    \label{fig:bc_vision_results}
\end{figure}

\section{Objective functions comparison for \textsc{DAgger}}
\label{sec:qual_comp}

In Figure~\ref{fig:dagger_objective_00} and in Figure~\ref{fig:dagger_objective_03}, we provide ablations over different objectives for \textsc{DAgger} with $\beta=0.0$ and $\beta=0.3$ correspondingly. We see that overall, training with cross-entropy leads to better results than with log prob on the mean action, especially when $\beta=0.0$.

\begin{figure}[h!]
    \includegraphics[width=0.9\linewidth]{images/bc-dagger-objective_00.pdf}
    \caption{\textbf{\textsc{DAgger} objective sweep with $\beta=0.0$.} On the X-axis we report the number of environment steps. On the Y-axis we report averaged across 3 seeds episodic reward achieved by the student. Shaded area corresponds to confidence intervals. For a Run task, the confidence intervals are small, so they are not visible. In solid line we report the performance when training using the cross-entropy. In dashed line, we report the performance when training using log probability on the mean action from the expert. All the methods use mean action during evaluation. The black dashed line indicate average (among the same 150 independent evaluations) expert performance for the given expert noise level.} 
    \label{fig:dagger_objective_00}
\end{figure}

\begin{figure}[h!]
    \includegraphics[width=0.9\linewidth]{images/bc-dagger-objective_03.pdf}
    \caption{\textbf{\textsc{DAgger} objective sweep with $\beta=0.3$.} On the X-axis we report the number of environment steps. On the Y-axis we report averaged across 3 seeds episodic reward achieved by the student. Shaded area corresponds to confidence intervals. For a Run task, the confidence intervals are small, so they are not visible. In solid line we report the performance when training using the cross-entropy. In dashed line, we report the performance when training using log probability on the mean action from the expert. All the methods use mean action during evaluation. The black dashed line indicate average (among the same 150 independent evaluations) expert performance for the given expert noise level.} 
    \label{fig:dagger_objective_03}
\end{figure}

\bibliography{main}